\documentclass[11pt]{article}
\usepackage[preprint]{acl}
\usepackage{times}
\usepackage{latexsym}
\usepackage{amsmath}
\usepackage{amssymb}
\usepackage{booktabs}
\usepackage{array}
\usepackage{placeins}
\usepackage{xcolor}
\usepackage{url}
\usepackage[T1]{fontenc}
\usepackage[utf8]{inputenc}

\newcolumntype{L}[1]{>{\raggedright\arraybackslash}p{#1}}
\newcolumntype{R}[1]{>{\raggedleft\arraybackslash}p{#1}}

\newcommand{\framework}{PREFACE}
\newcommand{\method}{PREFACE-IA3}
\newcommand{\sharedmethod}{PREFACE-IA3-shared}
\newcommand{\smallmethod}{PREFACE-IA3-small}
\newcommand{\gatedmethod}{PREFACE-IA3-gated}
\newcommand{\spectralmethod}{PREFACE-Spectral}
\newcommand{\lorasidecar}{PREFACE-LoRA}
\newcommand{\dorasidecar}{PREFACE-DoRA}
\newcommand{\secondbest}[1]{\underline{#1}}
\newcommand{\iafull}{Infused Adapter by Inhibiting and Amplifying Inner Activations}
\definecolor{cpblue}{RGB}{55,92,170}
\definecolor{cpbluefill}{RGB}{238,243,255}
\definecolor{cpgreen}{RGB}{54,132,89}
\definecolor{cpgreenfill}{RGB}{239,249,242}
\definecolor{cpgray}{RGB}{92,99,112}
\definecolor{cpgrayfill}{RGB}{247,248,250}
\definecolor{cpgold}{RGB}{155,111,31}
\definecolor{cpgoldfill}{RGB}{255,248,232}

\newcommand{\pngfigure}[2]{%
\begingroup
\leavevmode\pdfximage width #1 {#2}%
\pdfrefximage\pdflastximage
\endgroup}

\title{\framework{}: Separating Shared Corrections from Prefix Conditioning in PEFT}

\author{Novanto Yudistira \and One Octadion}

\begin{document}
\maketitle

\begin{abstract}
Does input-conditioned parameter-efficient fine-tuning (PEFT) improve because it uses the current input, or because it adds a useful learned correction?
We investigate this distinction with \framework{}, a bounded channel controller for PEFT-adapted projections that requires no test-time gradients.
The controller reads observed context; in multiple-choice likelihood scoring, its state is independent of the candidate continuation.
Our central study freezes IA3, tuned LoRA, and DoRA adapters and matches controller architecture, initialization, batches, and updates across current-prefix and learned-global controls.
Across ARC-Challenge and SciQ, current-prefix control improves all six mean accuracies over frozen adapters, but none of its paired contrasts with learned-global control excludes zero.
Replacing its corrections with their training-prefix mean changes accuracy by at most 0.34 percentage points; input-varying corrections contribute only 1.8--2.8\% of correction RMS.
On ARC test, frozen-adapter current/global controls score 25.68\%/25.77\%, distinct from the earlier jointly trained 23.1\% to 26.4\% gain over IA3.
We derive and evaluate static compilation of fixed corrections, separating deployment simplification from statistical equivalence.
Broader GPT-2, Pythia, and Qwen experiments show family-, task-, and scoring-dependent gains, with modest recent-model effects.
The study provides an intervention-based evaluation of conditional PEFT: improvements over a static adapter do not by themselves demonstrate a benefit from input matching.
\end{abstract}

\begin{figure*}[t]
\centering
\pngfigure{\textwidth}{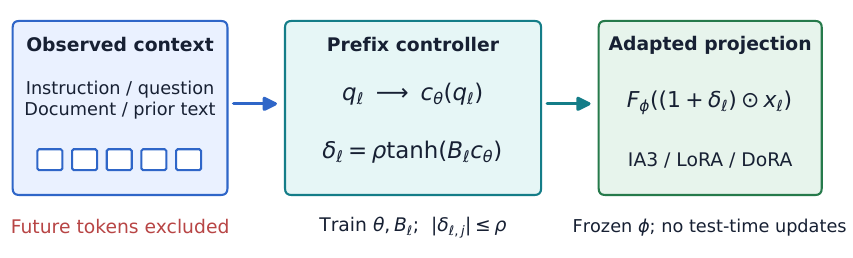}
\caption{The frozen-adapter PREFACE wrapper. Observed context is encoded into a bounded correction applied to the input of an adapted projection, with the backbone and base adapter fixed. The construction supports subsequent token- or sequence-level inference. Our MCQ protocol additionally reuses one context state across candidates. Earlier jointly trained multiplicative variants are distinguished in Section~\ref{sec:method}.}
\label{fig:method-overview}
\end{figure*}

\section{Introduction}

Pretrained language models are usually deployed with fixed weights.
Parameter-efficient fine-tuning (PEFT) methods such as LoRA \citep{hu2022lora} and \iafull{} (IA3; \citealp{liu2022fewshot}) freeze the backbone and train compact adapters, but the learned adapters are typically static after training.
Recent inference-time adaptation methods introduce more dynamic behavior.
Transformer-Squared modulates singular components through task dispatch and expert vectors \citep{sun2025transformersquared}, and In-Place Test-Time Training updates MLP fast weights with a next-token-prediction objective during inference \citep{feng2026inplace}.
These methods suggest that MLP blocks are promising adaptation sites, but they also raise a practical question: how much context-dependent adaptation is possible without test-time gradient updates or large per-task states?

Our starting point is an attribution problem: comparing an input-conditioned adapter only with its static predecessor changes both parameterization and input dependence.
Input-dependent strength might avoid under-correcting some inputs and over-correcting others, but a shared correction learned by the larger parameterization could also explain the gain.
We separate these explanations with trained global controls and interventions on a fixed trained controller.
\framework{} learns a small bounded change in channel strength from the observed prefix, either jointly with the static scale or around a frozen adapted projection.
The residual parameterization exactly recovers the static adapter when the controller output is zero and prevents the controller from generating an unconstrained replacement adapter.

For example, a static IA3 adapter uses the same scale vector $a_0$ for every instruction, question, or preceding text context.
\framework{} compresses the observed context into $c(q)$ and uses $a(q)=a_0+\rho\tanh(Bc(q))$.
The resulting scale modulates subsequent inference; in our MCQ experiments it is reused while scoring all candidate answers, so candidates cannot influence it.
For LoRA or DoRA, the evaluated wrapper rescales the input to the adapted projection rather than synthesizing a new low-rank matrix.
This yields two distinct questions: does adding the correction help, and does assigning it to the matching input help beyond a shared correction?
The second question is central to our study, not an assumption implicit in the method's name.

Our contributions are:
\begin{itemize}
    \item We define a common, candidate-independent correction interface and a matched IA3/LoRA/DoRA study separating added correction, input matching, and boundedness. Frozen/global/current comparisons and same-checkpoint interventions are the primary evidence.
    \item We quantify the difference between a variable controller representation and a functionally variable correction. Fixed-mean replacement and prediction transitions show why internal-state variation alone is insufficient evidence for useful input adaptation.
    \item We derive static compilation of fixed corrections and evaluate its numerical fidelity and runtime. Broader backbone, scoring, and objective diagnostics delimit the findings rather than imply universal transfer.
\end{itemize}

Figure~\ref{fig:method-overview} gives a plain-language view of the shared prefix controller used in our experiments.

\section{Related Work}

\paragraph{Parameter-efficient adaptation.}
PEFT methods differ in where they attach task-specific state.
Adapter modules insert small trainable bottlenecks between frozen Transformer components \citep{houlsby2019adapters}, prefix tuning and prompt tuning learn continuous conditioning vectors while keeping the backbone fixed \citep{li2021prefix,lester2021prompt}, and BitFit updates only bias terms \citep{benzaken2022bitfit}.
LoRA learns low-rank additive updates to frozen weights \citep{hu2022lora}.
Recent LoRA variants improve this additive-update family through SVD-based initialization or weight decomposition, including PiSSA \citep{meng2024pissa} and DoRA \citep{liu2024dora}.
IA3 instead learns multiplicative vectors that rescale inner activations, giving a very small parameter footprint \citep{liu2022fewshot}.
We use IA3 as a static multiplicative reference point, not as the definition of prefix-conditioned adaptation.
\sharedmethod{} instantiates the controller with residual channel scaling around the identity, while \lorasidecar{} and \dorasidecar{} test the same bounded-gating mechanism on tuned additive LoRA and DoRA adapters.

\paragraph{Input-conditioned and representation-side control.}
Several PEFT methods make adaptation depend on the input rather than using only a fixed adapter.
Hypernetworks generate one model's weights from another network's conditioning signal \citep{ha2017hypernetworks}.
Within PEFT, HyperAdapter uses task or example summaries to generate adapter parameters or mixtures \citep{zhao2023prototype}, while AdapterFusion composes trained adapters without changing their task-specific parameters \citep{pfeiffer2021adapterfusion}.
Representation finetuning methods such as ReFT/LoReFT intervene directly on hidden states rather than low-rank weight updates \citep{wu2024reft}.
Input-adaptive prompt methods such as ADePT \citep{tang2025adept} make prompt-side offsets depend on the input, while PARA generates prompt-aware representation adjustments inside Transformer layers for multi-tenant PEFT \citep{liu2024para}.
PEDRO similarly uses lightweight prompt-dependent vector generators to modify hidden representations through dot-product interactions \citep{xie2024pedro}.
\framework{} uses two design constraints to study this family of methods.
First, the controller state is computed only from context observed before downstream target tokens, enabling causal prefix-conditioned control for subsequent token- or sequence-level inference. For MCQ likelihood scoring, that state is reused unchanged for every candidate continuation, preventing the adaptive state from seeing answer-specific tokens.
Second, the controller changes only channel-wise gains inside MLP PEFT modules: the same prefix state can adjust IA3-style scales, tuned LoRA, tuned DoRA, spectral diagnostics, or compact representation-side interventions without changing the base adapter's internal parameterization.
This makes our intervention narrower than general representation editing while keeping the controller independent of a particular downstream scoring format and allowing a controlled test of transfer across PEFT families.
We do not claim a new class of hypernetwork or superiority over these methods: our main comparison asks whether the input dependence of our controller is necessary once the added correction and training protocol are controlled.

\paragraph{Spectral and test-time adaptation.}
SVFT and SVFit adapt pretrained singular directions or values \citep{lingam2024svft,sun2024svfit}; Transformer-Squared uses spectral expert dispatch \citep{sun2025transformersquared}.
Test-time training updates model states using inference-time self-supervision \citep{sun2020ttt,feng2026inplace}.
PREFACE uses a trained feed-forward controller with no inference-time gradients.
MLP key-value interpretations \citep{geva2021transformerffn} motivate our intervention site, but do not establish its performance mechanism.
Our compact LoReFT- and PARA/PEDRO-style references share the local evaluation pipeline; they are not official-code reproductions.
Appendix~\ref{app:implementation} summarizes the design differences and spectral implementation.

\section{Method}
\label{sec:method}

PREFACE learns bounded channel corrections from an observed context and applies them to a PEFT-adapted computation.
The construction supports subsequent token- or sequence-level inference.
We distinguish the controller's input dependence, which follows from its definition, from whether matching a correction to a particular input improves performance, which requires an intervention experiment.

\subsection{Bounded Controller}

Let $q_\ell$ summarize the observed context at layer $\ell$.
A shared small network produces coefficients $c_\theta(q_\ell)$, and a layer-specific basis $B_\ell$ maps them to a channel correction:
\begin{align}
c_\theta(q_\ell) &= W_2\,\mathrm{GELU}(W_1\,\mathrm{LN}(q_\ell)+b_1)+b_2,\\
\delta_\ell(q_\ell) &= \rho\tanh(B_\ell c_\theta(q_\ell)).
\label{eq:bounded-control}
\end{align}
Here $q_\ell$ is a mean of context-token activations, $c_\theta(q_\ell)\in\mathbb{R}^m$, $B_\ell\in\mathbb{R}^{d_\ell\times m}$, and $\rho>0$ is a fixed bound.
Thus each coordinate satisfies $|\delta_{\ell,j}|\leq\rho$.
The final controller layer is initialized to zero, recovering the static computation at initialization.
The encoder and bases are learned on training examples; inference uses a forward pass without parameter updates.
The same architecture also defines a learned-global control by replacing $q_\ell$ with a constant, and an unbounded control by removing $\tanh$.

\paragraph{Layer-wise order, without a cycle.}
Let $x_{\ell,t}$ be the already-computed input to the adapted projection at token $t$, before its current-layer correction, and let $P$ index observed-context tokens.
We first compute $q_\ell=|P|^{-1}\sum_{t\in P}x_{\ell,t}$, then $\delta_\ell(q_\ell)$, and finally the corrected projection in Eq.~\ref{eq:frozen-wrapper}.
Earlier layers' controls can affect $x_\ell$, but the control at layer $\ell$ is computed only after $x_\ell$ is available.
There is no fixed-point iteration or dependence on that layer's future output; sharing controller weights across layers adds no backward computational edge.

\subsection{Application Across Adapter Families}

Write $F_{\phi,\ell}$ for a projection containing a trained PEFT adapter.
Our frozen-adapter input wrapper computes
\begin{equation}
y_\ell=F_{\phi,\ell}\big((\mathbf{1}+\delta_\ell(q_\ell))\odot x_\ell\big).
\label{eq:frozen-wrapper}
\end{equation}
All parameters $\phi$ remain fixed.
For an affine adapted projection $F_{\phi,\ell}(x)=W_{\phi,\ell}x+b_\ell$, its local change at a fixed activation is exactly
\begin{equation}
\Delta y_\ell=W_{\phi,\ell}(\delta_\ell\odot x_\ell).
\label{eq:local-effect}
\end{equation}
Consequently, equal correction norms need not imply equal output changes: the activation and adapted projection also matter.
We measure these changes rather than assuming that a particular adapter family must benefit.
The coordinate bound implies $\|\Delta y_\ell\|_2\leq\rho\|W_{\phi,\ell}\|_2\|x_\ell\|_2$, not a guarantee of improved predictions.
If one fixed correction $\bar\delta_\ell$ suffices, it can be folded into $W_{\phi,\ell}\mathrm{diag}(\mathbf{1}+\bar\delta_\ell)$ without a dynamic controller; our fixed-mean intervention tests the accuracy consequence of this simplification.

For LoRA and DoRA, the evaluated input wrapper rescales the input to the complete adapted projection.
It does not rescale only the additive low-rank branch.
Its frozen weight is $W+\Delta W$ for LoRA and the magnitude-normalized adapted weight for DoRA.
A learned input-independent scale is additionally present in the earlier LoRA/DoRA runs.
The matched study removes this extra scale and uses Eq.~\ref{eq:frozen-wrapper} identically across families.

The earlier multiplicative sweeps instead use
\begin{equation}
y_\ell=W_\ell\big((a_{0,\ell}+\delta_\ell(q_\ell))\odot x_\ell\big)+b_\ell.
\end{equation}
Their static IA3 scale $a_{0,\ell}$ is jointly optimized in the main DistilGPT-2 runs and frozen in the two-stage gated variants.
These runs establish empirical performance but do not alone isolate adapter family from training protocol.
The new matched study freezes every base adapter and uses the same controller architecture, training schedule, and intervention site.

Representation-side instances rescale low-rank intervention values, while the spectral diagnostic rescales singular channels.
These are alternative intervention sites, not evidence that every site performs equally well.
Architecture-specific target layers, sharing variants, parameter counts, and the spectral equations appear in Appendix~\ref{app:implementation}.

\paragraph{Shared and input-varying components.}
Let $\bar\delta_\ell$ be the mean post-nonlinearity correction over training prefixes and let $\epsilon_\ell(p)=\delta_\ell(p)-\bar\delta_\ell$.
At a fixed incoming activation $x$, replacing current control by its mean changes the affine projection by exactly $W_{\phi,\ell}(\epsilon_\ell(p)\odot x)$.
This separates the shared correction from its input-varying residual; it is not a bound on final-network accuracy because changes propagate to later layers.
For a fixed correction, compilation is exact in real arithmetic:
\begin{equation}
\widetilde W_{\phi,\ell}=W_{\phi,\ell}\operatorname{diag}(\mathbf 1+\bar\delta_\ell).
\label{eq:compilation}
\end{equation}
Bias is unchanged. We test numerical fidelity of this compiled model separately from whether replacing dynamic control changes predictions.

\section{Experimental Setup}
\label{sec:setup}

\paragraph{Models and data.}
We evaluate DistilGPT-2, GPT-2 small/medium, Pythia-410M, Qwen2.5-0.5B/1.5B-Instruct, and Qwen3.5-0.8B \citep{qwen2026qwen35} within an 8GB GPU budget.
Tasks are SciQ \citep{welbl2017sciq}, ARC-Challenge \citep{clark2018arc}, WinoGrande \citep{sakaguchi2021winogrande}, and PIQA \citep{bisk2020piqa}.
Most results use full labeled validation splits.
One fixed DistilGPT-2 IA3/PREFACE pair is additionally evaluated on the untouched 1,172-example ARC-Challenge test split.
Validation-selected results, including the matched study below, are development diagnostics rather than independent confirmatory estimates.
Uniform guessing over each question's actual choice count gives 25.07\% on ARC validation and 25.02\% on ARC test (some questions have three or five choices), 25\% on SciQ, and 50\% on PIQA/WinoGrande.

\paragraph{Scoring and objective.}
Context-prefix (CP) scoring appends each answer text to \texttt{Question: \{question\}\textbackslash nAnswer:}; WinoGrande instead appends each option and sentence suffix to the prefix before the blank.
We report summed and length-normalized continuation log-probability.
First-token prediction with prefilling (FTP+prefill) scores the first discriminative answer token after any shared answer prefix, with choices listed in the prompt.
Multiple-choice prompting (MCP) lists all options and scores either the answer label or answer text.
Appendix Table~\ref{tab:protocol-definitions} gives exact definitions.
For input-conditioned scoring, only context activations enter the controller; its state is identical across candidates, with no answer-token pooling.
Training minimizes cross-entropy over candidate mean log-probabilities.
A separate SciQ teacher-forced answer-token negative log-likelihood (NLL) experiment tests a standard token objective beyond candidate ranking.

\paragraph{Reference experiments.}
Static references include IA3, tuned LoRA/DoRA, prompt/prefix tuning, and compact local LoReFT- and PARA/PEDRO-style implementations.
Headline GPT-style runs use 500 updates, batch size four, and maximum training length 96.
IA3 and its jointly trained controller use learning rate $3\times10^{-3}$.
LoRA ranks and learning rates are selected on validation; DoRA uses rank eight and $10^{-3}$.
Qwen3.5 uses one fixed configuration across three tasks: 1,024 training examples, 200 updates, and three seeds.
Most studies use seeds 13/21/42; GPT-2 small ARC uses five.
Appendix~\ref{app:implementation} preserves all sweep settings, prompts, target layers, and variant details.

\paragraph{Matched frozen-adapter study.}
To isolate the effect of input-dependent corrections from protocol differences, we freeze existing IA3, tuned LoRA, and DoRA checkpoints on DistilGPT-2 and apply Eq.~\ref{eq:frozen-wrapper} identically.
For each family, task (ARC/SciQ), and seed (13/21/42), we train three arms: current-prefix bounded control; learned-global bounded control with constant $q_\ell$; and current-prefix unbounded control.
All arms have the same 129,108 nominal trainable parameters, initial weights, sampled training batches, and 200-update budget, without a separately trainable static scale. Equal nominal counts do not imply equal input-dependent expressivity or identical gradients when the input is constant.
The shared controller has hidden width 16, four coefficients, and $\rho=0.2$.
We use 1,024 training examples, batch size four, AdamW at $5\times10^{-4}$, weight decay 0.01, clipping 1.0, and dropout disabled.
The protocol is fixed before examining these runs; existing base-adapter selection still limits independence.
No newly observed outcome selects a configuration.

\paragraph{Interventions and uncertainty.}
On each bounded current-prefix checkpoint, we evaluate the current correction, zero correction (exact frozen-adapter recovery), training-mean correction, random correction, and three shuffled donor permutations.
The mean and random-scale estimates use 256 training prefixes, never validation prefixes.
Random corrections are norm-matched around this mean and clipped to the coordinate bound; they do not preserve learned covariance.
Shuffles reuse complete correction vectors from other validation examples, with no self-donors or labels.
All choices use the same assigned vectors.
For the matched study, paired crossed bootstrap resamples seeds and common example IDs (200,000 replicates); permutations are averaged within seeds, not treated as independent trials.
We distinguish pointwise 95\% intervals from Bonferroni sensitivity intervals over 12 comparisons (six task--family cells against both frozen and global controls), and over all 36 matched-study contrasts.
The additional ARC test contrast has a separate 37-comparison sensitivity analysis (Appendix~\ref{app:ac-statistics}). These adjustments do not cover every comparison in the paper or remove validation-selection bias.
Earlier results retain their reported, pointwise hierarchical intervals. All estimates remain exploratory with few seeds; pooled McNemar is only an appendix diagnostic.

\paragraph{Compilation audit.}
We compile the saved training-mean corrections for all 18 bounded-current checkpoints, without training or selecting a new setting, and check all answer scores on the full validation splits against the uncompiled fixed-mean intervention.
Runtime uses the first 24 preserved examples per task, seed 13, one question per batch with all candidates, float32, four warmup questions and five interleaved timing blocks on an RTX 5060 Laptop GPU.
Dynamic timing includes a prefix-only pass and candidate scoring without KV-cache reuse; compiled timing requires neither a controller nor a prefix-only pass.
CUDA-synchronized timing excludes loading and tokenization. This is a matched implementation benchmark, not an optimal implementation comparison; full protocol and memory definitions appear in Appendix~\ref{app:compilation}.

\section{Results}
\label{sec:main-results}

\subsection{Matched Mechanism Controls}
\label{sec:matched-study}

\begin{table*}[t]\centering\small
\setlength{\tabcolsep}{4pt}
\begin{tabular}{@{}llrrrrrr@{}}\toprule
Task & Adapter & Frozen & Global & Current & Fixed mean & Random & Shuffled \\
\midrule
ARC-C & IA3 & 24.4$\pm$0.3 & 27.6$\pm$1.9 & 28.2$\pm$0.8 & \textbf{28.5$\pm$0.7} & 28.4$\pm$0.7 & 28.3$\pm$0.7 \\
ARC-C & LoRA & 32.6$\pm$2.7 & 33.9$\pm$1.6 & 34.1$\pm$1.2 & 34.2$\pm$1.4 & 34.1$\pm$1.2 & \textbf{34.3$\pm$1.2} \\
ARC-C & DoRA & 34.3$\pm$4.0 & 35.1$\pm$1.2 & 35.9$\pm$1.5 & \textbf{36.0$\pm$1.5} & \textbf{36.0$\pm$1.4} & 35.9$\pm$1.5 \\
\midrule
SciQ & IA3 & 51.5$\pm$0.9 & \textbf{53.0$\pm$0.6} & 52.9$\pm$0.6 & 52.9$\pm$0.6 & \textbf{53.0$\pm$0.7} & \textbf{53.0$\pm$0.7} \\
SciQ & LoRA & 51.4$\pm$1.6 & 52.5$\pm$1.0 & 52.7$\pm$1.0 & \textbf{52.8$\pm$0.9} & \textbf{52.8$\pm$0.9} & \textbf{52.8$\pm$0.9} \\
SciQ & DoRA & 50.6$\pm$0.8 & \textbf{53.6$\pm$0.6} & 53.5$\pm$0.6 & 53.3$\pm$0.6 & 53.3$\pm$0.6 & 53.3$\pm$0.6 \\
\bottomrule\end{tabular}
\caption{Matched frozen-adapter study: DistilGPT-2, full validation splits (299 ARC-C; 1,000 SciQ), three seeds. Uniform guessing gives 25.07\%/25\%, respectively. Current/global each add 129,108 nominal parameters with matched initialization, batches and 200 updates. Global receives zero input; mean/random/shuffled intervene on the current checkpoint without retraining. Shuffled averages three permutations per seed. Bold marks the highest displayed mean, including rounding ties, not significance.}
\label{tab:matched-mechanism}
\end{table*}

\begin{table*}[t]\centering\small
\setlength{\tabcolsep}{4pt}
\begin{tabular}{@{}llrrr@{}}\toprule
Task & Adapter & Current $-$ frozen [95\%] & Current $-$ global [95\%] & Global contrast [B12] \\
\midrule
ARC-C & IA3 & +3.79 [0.00, 7.47] & +0.56 [-1.45, 3.01] & [-2.34, 4.35] \\
ARC-C & LoRA & +1.56 [-1.78, 4.91] & +0.22 [-1.67, 2.34] & [-2.68, 3.68] \\
ARC-C & DoRA & +1.56 [-2.34, 5.91] & +0.78 [-1.67, 2.79] & [-3.46, 3.90] \\
SciQ & IA3 & +1.40 [-0.80, 3.60] & -0.03 [-0.87, 0.80] & [-1.30, 1.23] \\
SciQ & LoRA & +1.33 [-0.70, 3.30] & +0.23 [-0.60, 1.10] & [-1.07, 1.67] \\
SciQ & DoRA & +2.87 [1.00, 4.77] & -0.10 [-0.90, 0.67] & [-1.37, 1.10] \\
\midrule
\multicolumn{5}{@{}l}{\textit{Additional frozen-IA3 test diagnostic: 1,172 ARC examples, three seeds}} \\
\multicolumn{5}{@{}l}{Frozen: 23.07$\pm$0.40; global: 25.77$\pm$0.45; current: 25.68$\pm$0.98.} \\
\multicolumn{5}{@{}l}{Current $-$ global: -0.09 [-1.25, 0.91]; B37: [-2.30, 1.54].} \\
\bottomrule\end{tabular}
\caption{Paired conditioning residuals in percentage points. [95\%] denotes pointwise crossed-bootstrap intervals (200,000 draws); B12 targets 95\% familywise coverage over six current--frozen and six current--global contrasts. The added test comparison uses existing frozen-adapter checkpoints without retraining or test-driven selection; B37 includes the 36 earlier matched-study contrasts. It is distinct from the jointly trained 26.4\% headline test result. No current--global interval excludes zero; this is not an equivalence claim.}
\label{tab:matched-other}
\end{table*}

Table~\ref{tab:matched-mechanism} separates three questions: whether adding the controller helps a frozen adapter; whether the learned correction must be matched to the current input; and whether independently trained global control is sufficient.
Unlike the earlier sweeps, all families use the same frozen-projection wrapper and controller budget.

\paragraph{Transfer across families.}
Current-prefix control improves the mean over the frozen adapter in all six task--family cells.
ARC gains are +3.79, +1.56, and +1.56 points for IA3, LoRA, and DoRA; the corresponding SciQ gains are +1.40, +1.33, and +2.87.
Only SciQ--DoRA excludes zero pointwise: +2.87 [1.00, 4.77], improving 50.6\% to 53.5\%; its B12 interval [0.00, 5.83] and B36 interval [-0.40, 6.30] do not exclude zero.
This is exploratory evidence beyond IA3/ARC, not a familywise-significant or universal advantage.
These development results do not replace the independent ARC test result below.

\paragraph{Does the current prefix matter?}
Replacing each input's correction by the training-mean correction changes accuracy by at most 0.34 points in any cell, with all current-minus-mean intervals including or touching zero.
Current-prefix control also does not reliably outperform shuffled or learned-global control in any cell.
Table~\ref{tab:matched-other} directly reports this conditioning residual per family and task, and extends current versus global to ARC test: 25.68\% versus 25.77\%, or -0.09 [-1.25, 0.91] points.
These are not equivalence tests, but the observed current--mean differences are small, and prediction disagreement is only 0.11--0.56\% across the six cells.
The results support a largely input-independent correction in this regime, not a demonstrated semantic-matching advantage.
They also motivate a concrete deployment test: the fixed-mean correction can be folded into the frozen projection weights (Eq.~\ref{eq:compilation}).
Section~\ref{sec:compilation} tests this compilation, without treating a small dynamic--mean difference as statistical equivalence.

\begin{figure*}[t]
\centering
\pngfigure{\textwidth}{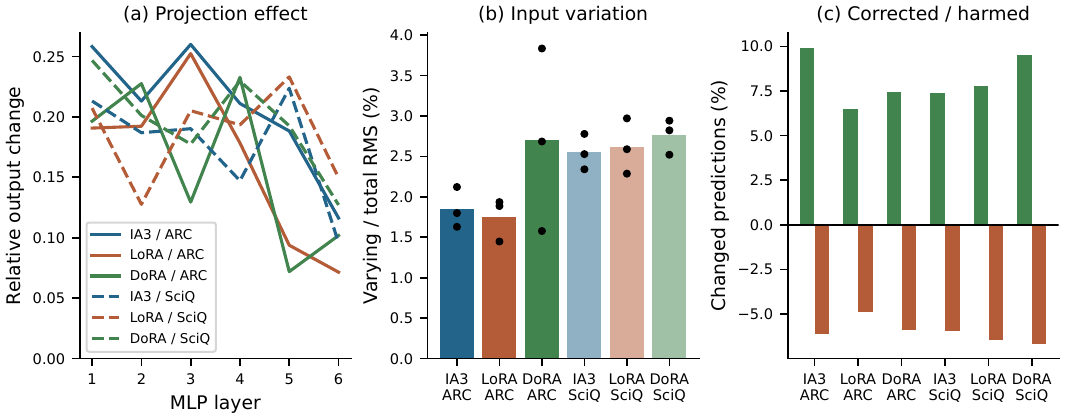}
\caption{Direct correction diagnostics in the matched frozen-adapter study. (a) Relative local output change at each MLP projection, holding its incoming activation fixed; solid lines denote ARC and dashed lines SciQ. This is not an attribution of the final prediction to one layer. (b) Input-varying correction RMS around the training mean, relative to total correction RMS; dots are seeds. (c) Percentages of predictions corrected and harmed relative to the frozen base adapter. A large correction is not necessarily useful: accuracy gain is the difference between corrected and harmed examples.}
\label{fig:correction-mechanism}
\end{figure*}

\paragraph{Which corrections are useful?}
Figure~\ref{fig:correction-mechanism} examines the output correction rather than merely probing the controller's internal state.
The input-varying residual around the training mean is only 1.8--2.8\% of total correction RMS, averaged over layers and seeds within each task--family cell.
A non-collapsed prefix representation therefore need not produce materially different correction vectors.
Output perturbations vary across layers and families, as Eq.~\ref{eq:local-effect} predicts; their magnitude alone does not explain accuracy.
On ARC, IA3 corrects 9.92\% of examples while harming 6.13\%, compared with LoRA's 6.47\%/4.91\% and DoRA's 7.47\%/5.91\%.
On SciQ, DoRA instead has the strongest net trade-off, 9.53\% corrected versus 6.67\% harmed.
The reversal across tasks argues against a universal IA3-only explanation. Adapted projections, activation statistics, and decision margins can change the effect of the same correction, as Eq.~\ref{eq:local-effect} makes explicit; these diagnostics localize the prediction trade-offs but do not identify which factor causes them.

\paragraph{Is the bound necessary?}
The equally initialized unbounded arm has no reliably different accuracy in any cell (Appendix Table~\ref{tab:ac-unbounded}).
The coordinate bound is a design constraint and yields the local norm inequality in Section~\ref{sec:method}; these results do not establish it as the cause of the accuracy gains.
Random-control clipping is zero in all runs, and its near-current accuracy cannot be attributed to clipping away the perturbation.
All accuracy, NLL, margin, and intervention outcomes are retained in the accompanying evidence package.

\subsection{Static Compilation}
\label{sec:compilation}
Table~\ref{tab:folding} tests a deployment consequence of the fixed-mean intervention, not a newly trained adapter. Compilation reproduces the uncompiled fixed-mean prediction on 11,690 of 11,691 seed--example pairs across all 18 checkpoints. The largest absolute answer-score discrepancy is $3.82\times10^{-5}$; one near-tied SciQ--IA3 prediction changes under floating-point reassociation without changing correctness. All 18 full-split accuracies are unchanged.
In this prefix-reuse implementation, compiled latency is 2.16--2.25 times lower than dynamic latency and close to the merged frozen adapter. These ratios include the dynamic prefix-only pass, but no KV-cache optimization; they should not be read as a hardware-independent speedup. Compilation removes the need to keep or evaluate the controller, but materializing a separately adapted dense model does not reduce backbone storage or guarantee a multi-tenant memory benefit.
\begin{table*}[t]\centering\small
\setlength{\tabcolsep}{4.5pt}
\begin{tabular}{@{}llrrrrrr@{}}\toprule
 & & \multicolumn{3}{c}{Accuracy (\%, three seeds)} & \multicolumn{3}{c}{Latency (ms/question)} \\
Task & Adapter & Current & Fixed mean & Compiled & Frozen & Current & Compiled \\\midrule
ARC-C & IA3 & 28.21 & 28.54 & 28.54 & 8.57 & 18.90 & \textbf{8.43} \\
ARC-C & LoRA & 34.11 & 34.23 & 34.23 & 8.47 & 18.29 & \textbf{8.48} \\
ARC-C & DoRA & 35.90 & 36.01 & 36.01 & 8.47 & 18.30 & \textbf{8.44} \\
\midrule
SciQ & IA3 & 52.93 & 52.93 & 52.93 & 8.14 & 18.16 & \textbf{8.19} \\
SciQ & LoRA & 52.73 & 52.80 & 52.80 & 8.17 & 18.23 & \textbf{8.19} \\
SciQ & DoRA & 53.50 & 53.33 & 53.33 & 8.15 & 18.25 & \textbf{8.13} \\
\bottomrule\end{tabular}
\caption{Static compilation using the training-prefix mean. Accuracy uses full validation splits; fixed-mean values come from the same checkpoints as Table~\ref{tab:matched-mechanism}. Latency is the median of five blocks on 24 questions per task, seed 13, RTX 5060 Laptop GPU, float32, batch size one question with all choices. Bold identifies the lower latency among current/compiled deployment modes, not accuracy significance. This is not a comparison of differently trained methods.}
\label{tab:folding}
\end{table*}

\subsection{Broader Empirical Evidence}

Appendix Table~\ref{tab:main-effect} preserves earlier same-family pairs: IA3 versus \smallmethod{}, tuned LoRA versus \lorasidecar{}, and tuned DoRA versus \dorasidecar{}.
These are within-family comparisons; the matched frozen-adapter study in Section~\ref{sec:matched-study} additionally controls training protocol.
The strongest pattern is on ARC-Challenge, where multiplicative prefix control improves over static IA3 on DistilGPT-2, GPT-2 small, and Pythia-410M.
The additive-family result is mixed: \lorasidecar{} improves DistilGPT-2 separation-style ARC-Challenge from 32.6\% to 34.3\%, while \dorasidecar{} is neutral under length-normalized scoring.
Appendix Table~\ref{tab:scoring-protocols} compares CP text scoring with first-token prefilling; mixed MCP stress tests appear in Appendix Table~\ref{tab:mcp-stress-test}.
Appendix Table~\ref{tab:prompt-side-baselines} compares learned virtual-token methods, while near-chance WinoGrande results remain a stability diagnostic.

Prompt tuning is smaller than \smallmethod{} but weaker on both datasets; prefix tuning wins on ARC-Challenge with nearly nine times as many parameters and a large SciQ regression.
Thus \framework{} is a compact controller over existing PEFT adapters, not a uniformly dominant replacement for prompt-side PEFT.

\begin{table}[!htbp]
\centering
\scriptsize
\setlength{\tabcolsep}{2.2pt}
\begin{tabular}{@{}L{1.65cm}rR{1.02cm}R{1.08cm}L{2.05cm}@{}}
\toprule
Evaluation & $n$ & IA3 & PREFACE & Paired $\Delta$ [95\%] \\
\midrule
DistilGPT-2 / ARC test & 1,172 & 23.1$\pm$0.4 & \textbf{26.4$\pm$1.4} & \textbf{+3.36 [1.39, 5.26]} \\
Qwen3.5 / ARC val. & 299 & 47.4$\pm$0.7 & \textbf{47.9$\pm$0.4} & +0.56 [-0.56, 1.67] \\
Qwen3.5 / SciQ val. & 1,000 & 84.7$\pm$0.8 & \textbf{85.1$\pm$0.9} & \textbf{+0.40 [0.03, 0.80]} \\
Qwen3.5 / PIQA val. & 1,838 & 70.2$\pm$0.4 & \textbf{70.4$\pm$0.5} & +0.15 [-0.20, 0.47] \\
\bottomrule
\end{tabular}
\caption{Held-out evaluation and recent-model transfer, with length-normalized scoring and three seeds. The ARC row evaluates the jointly trained -small checkpoint against IA3 on the test split withheld from its selection (uniform guessing: 25.02\%); it is not the matched frozen-adapter test in Table~\ref{tab:matched-other}. Qwen3.5-0.8B uses a fixed configuration and full validation splits. Intervals are pointwise hierarchical estimates.}
\label{tab:qwen-clean-main}
\end{table}

Fixed DistilGPT-2 checkpoints retain the ARC-Challenge gain on the untouched test split in all seeds.
With one Qwen3.5 configuration across three datasets, all mean deltas are positive but only SciQ excludes zero; the macro average rises from 67.45 to 67.82.

\smallmethod{} also lowers SciQ answer-token NLL relative to IA3 for all three seeds, suggesting that bounded multiplicative control can help outside answer-choice ranking, though we keep this as a scoped evaluation rather than a broad generation benchmark.

\paragraph{Additional diagnostics.}
Appendix diagnostics report full SciQ accuracy, Qwen SciQ, MCP-style scoring stress tests, prefix-source diagnostics, controller ablations, boundedness, paired evidence, and latency; together they support the ARC-Challenge multiplicative result while preserving the mixed-task interpretation.

\section{Conclusion}
\label{sec:conclusion}

We studied \framework{}, a prefix-conditioned controller that makes small bounded changes around static PEFT adapters for token- or sequence-level inference without test-time gradients; causal prefix reuse additionally prevents candidate-answer leakage in our MCQ protocol.
The controller can modify IA3-adapted projections, LoRA/DoRA-adapted projections, and representation-side low-rank channels.
Earlier cross-backbone evidence is strongest for IA3 on ARC-Challenge; the matched study also has a positive SciQ--DoRA point estimate, but its multiplicity-adjusted interval touches zero.
Fixed-mean and learned-global controls largely retain the gains, including in the ARC test diagnostic, and bounded versus unbounded differences remain uncertain.
The contribution is a controlled study separating added correction from input matching, with a tested static-compilation route when the input-varying component is dispensable in the evaluated setting.
The results do not establish statistical equivalence, a universal failure of conditional PEFT, or boundedness as the cause of the gains.

\FloatBarrier
\section{Limitations}
\label{sec:limitations}

\paragraph{Limitations.}
The evidence is concentrated on small GPT-2/GPT-NeoX models plus Qwen2.5 and Qwen3.5-0.8B, mostly MCQ likelihood evaluation and one SciQ answer-token likelihood evaluation. The untouched ARC-Challenge test result reduces validation-selection risk for the main DistilGPT-2 claim, but most other rows remain validation diagnostics. Results are strongest on ARC-Challenge and mixed on SciQ, WinoGrande, PIQA, GPT-2 medium, additive PEFT, DoRA, and MCP-style option prompting.
The prompt-side baseline comparison is also limited: prefix tuning outperforms \smallmethod{} on DistilGPT-2 ARC-Challenge at much higher parameter cost, while prompt tuning and SciQ prefix tuning are weaker under our loop.
Full external-code LoReFT/PARA/PEDRO sweeps, exhaustive prompt/prefix hyperparameter searches, larger modern backbones, broader generation/classification tasks, and controller compression remain open.
Three-seed matched diagnostics and validation-selected base settings limit inference even when evaluation uses complete splits. Test results already examined here cannot serve as an untouched selection-independent test for future modifications. Static compilation validates this implementation on the reported data and precision, not all future inputs, hardware, decoding modes, or conditional-PEFT methods.

\section*{Ethical Considerations}
We introduce no new dataset of people or deployed application.
Input-conditioned adapters may amplify context-specific biases, so we report regressions and avoid factuality or safety conclusions without direct evaluation.

\section*{Acknowledgements}
AI assistance was used for research-idea exploration, code development and debugging, analysis scripts, manuscript drafting and revision, and preparation of submission materials. Responsibility for the research, verification, and final submission remains with the human authors.

\paragraph{Code and numerical artifacts.}
The accompanying arXiv ancillary archive, \texttt{PREFACE\_code.zip}, contains implementation code, evaluation outputs, and instructions for recomputing the reported analyses. Pretrained model weights and training datasets are obtained separately under their respective licenses.

\bibliography{references}

\clearpage
\appendix
\section{Earlier Same-Family and Scoring Comparisons}
\label{app:earlier-results}
These previously reported comparisons preserve the original training protocols and separate their adapter gains from the matched input-conditioning tests in the main paper.
\begin{table*}[t]
\centering
\scriptsize
\setlength{\tabcolsep}{2.3pt}
\begin{tabular}{@{}llp{3.1cm}rrrrl@{}}
\toprule
PEFT family & Setting & Same-family comparison & Params & Seeds & Static & PREFACE & $\Delta$ \\
\midrule
\multicolumn{8}{@{}l}{\textit{Headline ARC-Challenge gains}} \\
IA3 & DistilGPT-2 / ARC-C & IA3 $\rightarrow$ \smallmethod{} & 147,540 & 3 & 24.4$\pm$0.3 & \textbf{28.7$\pm$2.8} & \textbf{+4.2} \\
IA3 & GPT-2 small / ARC-C & IA3 $\rightarrow$ \smallmethod{} & 239,700 & 5 & 24.5$\pm$1.5 & \textbf{29.0$\pm$2.3} & \textbf{+4.5} \\
IA3 & Pythia-410M / ARC-C & IA3 $\rightarrow$ \gatedmethod{} & 467,052 & 3 & 31.5$\pm$1.5 & \textbf{34.0$\pm$0.8} & \textbf{+2.5} \\
IA3 & Qwen2.5-1.5B / ARC-C & IA3 $\rightarrow$ \gatedmethod{} & 1,164,912 & 3 & 51.5$\pm$1.0 & \textbf{53.1$\pm$0.2} & \textbf{+1.6} \\
LoRA & DistilGPT-2 / ARC-C & LoRA-tuned $\rightarrow$ \lorasidecar{} & 270,632 & 3 & 32.6$\pm$2.7 & \textbf{34.3$\pm$0.7} & \textbf{+1.8} \\
DoRA & DistilGPT-2 / ARC-C & DoRA $\rightarrow$ \dorasidecar{} & 270,632 & 3 & 34.3$\pm$4.0 & \textbf{34.4$\pm$3.2} & +0.1 \\
\midrule
\multicolumn{8}{@{}l}{\textit{Stability and non-win diagnostics}} \\
IA3 & DistilGPT-2 / SciQ & IA3 $\rightarrow$ \smallmethod{} & 147,540 & 3 & 51.5$\pm$0.9 & \textbf{52.6$\pm$0.5} & \textbf{+1.1} \\
\multicolumn{8}{@{}l}{\textit{Answer-token likelihood evaluation}} \\
IA3 & DistilGPT-2 / SciQ answer NLL & IA3 $\rightarrow$ \smallmethod{} & 147,540 & 3 & 4.488$\pm$0.005 & \textbf{4.464$\pm$0.007} & \textbf{-0.024} \\
\bottomrule
\end{tabular}
\caption{Headline same-family comparisons, not parameter-matched conditioning tests. ARC uniform-guessing accuracy is 25.07\%. DistilGPT-2/GPT-2 IA3 rows use the jointly trained -small variant; Pythia/Qwen2.5-1.5B use the two-stage -gated variant. Params count trainable parameters in the reported adaptation stage, including the static scale when jointly trained. SciQ answer NLL is lower-is-better, with pointwise paired interval -0.024 [-0.040, -0.007].}
\label{tab:main-effect}
\end{table*}
\begin{table}[t]
\centering
\scriptsize
\setlength{\tabcolsep}{2.1pt}
\begin{tabular}{@{}llrrl@{}}
\toprule
PEFT family & Protocol & Static adapter & PREFACE & $\Delta$ \\
\midrule
IA3 & CP text & 24.4$\pm$0.3 & \textbf{28.7$\pm$2.8} & +4.2 [0.1, 8.1] \\
LoRA & CP text & 32.6$\pm$2.7 & \textbf{34.3$\pm$0.7} & +1.8 [-1.4, 5.1] \\
DoRA & CP text & 34.3$\pm$4.0 & \textbf{34.4$\pm$3.2} & +0.1 [-3.1, 3.5] \\
\midrule
IA3 & FTP+prefill & 19.0$\pm$0.8 & \textbf{23.0$\pm$2.3} & +4.0 [1.3, 6.8] \\
LoRA & FTP+prefill & 25.9$\pm$1.7 & \textbf{26.2$\pm$0.4} & +0.3 [-1.9, 2.6] \\
DoRA & FTP+prefill & 25.2$\pm$1.8 & \textbf{25.9$\pm$1.8} & +0.7 [-1.7, 3.0] \\
\bottomrule
\end{tabular}
\caption{ARC-Challenge protocol diagnostic on DistilGPT-2 using the same CP-trained checkpoints (uniform-guessing accuracy: 25.07\%). CP scores answer text; FTP+prefill lists choices and scores the first discriminative token after any shared prefix. Deltas use pointwise hierarchical intervals, not multiplicity-adjusted evidence. MCP stress tests appear in Appendix Table~\ref{tab:mcp-stress-test}.}
\label{tab:scoring-protocols}
\end{table}
\section{SciQ Accuracy Details}
\label{app:sciq-results}

Table~\ref{tab:sciq-results} reports the full SciQ validation accuracy table under both summed and length-normalized continuation scoring.
It complements the main-text SciQ likelihood result by showing that PREFACE-IA3 is competitive but only modestly better under the standard MCQ scoring setup.
The table also makes the metric dependence explicit: LoRA-tuned is strongest under summed continuation likelihood, whereas \smallmethod{} is strongest under length-normalized scoring.
This is why the paper treats SciQ as supportive rather than decisive evidence for the controller design.

\begin{table}[!htbp]
\centering
\scriptsize
\setlength{\tabcolsep}{3.6pt}
\begin{tabular}{lrrrr}
\toprule
Method & Trainable params & Seeds & Sum & Mean \\
\midrule
DistilGPT-2 & 0 & 3 & 40.3$\pm$0.1 & 40.5$\pm$0.1 \\
LoRA-r8 (initial) & 184,320 & 3 & 27.1$\pm$1.2 & 34.0$\pm$2.1 \\
LoRA-tuned & 368,640 & 3 & \textbf{43.5$\pm$0.7} & 51.4$\pm$1.6 \\
DoRA & 188,928 & 3 & 40.4$\pm$2.1 & 50.6$\pm$0.8 \\
\midrule
IA3 & 18,432 & 3 & 42.0$\pm$0.5 & 51.5$\pm$0.9 \\
\sharedmethod{} & 270,632 & 3 & 41.9$\pm$0.8 & \secondbest{52.4$\pm$0.9} \\
\smallmethod{} & 147,540 & 3 & \secondbest{42.5$\pm$0.9} & \textbf{52.6$\pm$0.5} \\
\spectralmethod{} & 4,802,358 & 3 & 41.0$\pm$0.3 & 41.8$\pm$0.2 \\
\bottomrule
\end{tabular}
\caption{SciQ accuracy on the full validation set, averaged over seeds 13, 21, and 42. ``Sum'' uses summed continuation log-probability; ``Mean'' uses length-normalized continuation log-probability. Bold marks the best score and underline marks the second-best score within each metric column. Input-conditioned rows use context-prefix-only controller states.}
\label{tab:sciq-results}
\end{table}

\section{Stress-Test Diagnostics}
\label{app:stress-tests}

Table~\ref{tab:qwen-sciq-stress} reports complementary Qwen2.5 SciQ evaluations.
Unlike the ARC-Challenge rows in Table~\ref{tab:qwen-clean-main}, these outcomes are mixed and the paired intervals overlap zero, so they serve as scope-calibration evidence rather than the core backbone-transfer result.
They help show that the modern-backbone effect is task-sensitive rather than uniformly positive across datasets.

\begin{table}[!htbp]
\centering
\scriptsize
\setlength{\tabcolsep}{4.0pt}
\begin{tabular}{@{}lrrl@{}}
\toprule
Base adapter & Static & PREFACE & Paired delta \\
\midrule
IA3 & \textbf{81.7$\pm$0.5} & 81.4$\pm$0.1 & -0.3 [-0.8, 0.2] \\
LoRA & \textbf{75.0$\pm$2.8} & 75.0$\pm$2.8 & 0.0 [-0.1, 0.0] \\
\bottomrule
\end{tabular}
\caption{Modern-backbone SciQ stress-test diagnostics on Qwen2.5-0.5B-Instruct. These rows are mixed or tied, so they are reported as complementary scope-calibration evidence alongside the ARC-Challenge backbone-transfer table.}
\label{tab:qwen-sciq-stress}
\end{table}

Table~\ref{tab:qwen35-full} reports both scoring rules for the Qwen3.5 study summarized in the main paper.
The same fixed hyperparameters are used for all three datasets.

\begin{table*}[!htbp]
\centering
\scriptsize
\setlength{\tabcolsep}{4.0pt}
\begin{tabular}{@{}llrrrrl@{}}
\toprule
Dataset & Score & $n$ & IA3 & PREFACE & $\Delta$ & Hierarchical 95\% interval \\
\midrule
ARC-Challenge & sum & 299 & 43.03$\pm$1.35 & \textbf{43.14$\pm$1.00} & +0.11 & [-0.67, 0.89] \\
ARC-Challenge & mean & 299 & 47.38$\pm$0.70 & \textbf{47.94$\pm$0.39} & +0.56 & [-0.56, 1.67] \\
SciQ & sum & 1,000 & 82.30$\pm$0.98 & \textbf{82.63$\pm$0.81} & +0.33 & [0.07, 0.67] \\
SciQ & mean & 1,000 & 84.73$\pm$0.76 & \textbf{85.13$\pm$0.85} & +0.40 & [0.03, 0.80] \\
PIQA & sum & 1,838 & 69.44$\pm$0.08 & \textbf{69.73$\pm$0.17} & +0.29 & [0.00, 0.58] \\
PIQA & mean & 1,838 & 70.24$\pm$0.44 & \textbf{70.38$\pm$0.54} & +0.15 & [-0.20, 0.47] \\
\bottomrule
\end{tabular}
\caption{Complete Qwen3.5-0.8B results over seeds 13, 21, and 42. Separate adapters are trained for each dataset using one fixed configuration. Positive effects are modest; SciQ excludes zero under both scoring rules, PIQA sum touches zero, and the remaining intervals overlap zero.}
\label{tab:qwen35-full}
\end{table*}

\section{Scoring Protocol Definitions}
\label{app:protocol-definitions}

Table~\ref{tab:protocol-definitions} makes the controller-visible and scored tokens explicit for every reported protocol.

\begin{table*}[!htbp]
\centering
\scriptsize
\setlength{\tabcolsep}{4.0pt}
\begin{tabular}{@{}lL{4.0cm}L{4.0cm}L{4.2cm}@{}}
\toprule
Protocol & Controller-visible prefix & Scored tokens & Shared-prefix handling \\
\midrule
CP text & Question/context only & Candidate answer text & None; one prefix representation is reused across choices. \\
FTP+prefill & Question plus visible options & First discriminative answer token & Tokens shared by all answer strings are prefixed but not scored. \\
MCP label & Question plus labeled options & Next option-label token & No answer-text continuation is exposed to the controller. \\
MCP text & Question plus labeled options & Candidate answer text & Each candidate text is masked from the controller. \\
Options-trained & Question plus labeled options & Candidate answer text & Training and evaluation use the same options-visible template. \\
\bottomrule
\end{tabular}
\caption{Scoring protocols. CP denotes context-prefix answer-text scoring, FTP denotes first-token prefilling, and MCP denotes multiple-choice prompting. Candidate tokens receiving likelihood scores never enter the controller representation.}
\label{tab:protocol-definitions}
\end{table*}

Table~\ref{tab:mcp-stress-test} gives the option-prompting stress tests referenced in Section~\ref{sec:main-results} and shows why CP text and first-token prefilling remain the main scoring protocols.
The label-prompt and full-text MCP formats change the ranking behavior enough that gains no longer line up cleanly across adapter families.
Reporting these rows separately keeps the main scoring table centered on the primary evaluation protocols while testing whether the same prefix-conditioned interface transfers across scoring formats.

\begin{table}[!htbp]
\centering
\scriptsize
\setlength{\tabcolsep}{3.2pt}
\begin{tabular}{@{}llrrl@{}}
\toprule
Base adapter & Protocol & Static adapter & PREFACE & $\Delta$ \\
\midrule
IA3 & MCP label & 21.3$\pm$0.2 & \textbf{21.4$\pm$0.0} & +0.1 [0.0, 0.4] \\
LoRA & MCP label & \textbf{24.3$\pm$0.5} & 23.2$\pm$2.6 & -1.1 [-4.9, 2.7] \\
DoRA & MCP label & \textbf{26.4$\pm$2.9} & 25.9$\pm$1.3 & -0.6 [-3.6, 2.2] \\
\midrule
IA3 & MCP text & \textbf{29.2$\pm$0.8} & 28.9$\pm$4.0 & -0.3 [-5.0, 4.7] \\
LoRA & MCP text & 31.8$\pm$4.1 & \textbf{33.0$\pm$3.9} & +1.2 [-1.3, 3.8] \\
DoRA & MCP text & \textbf{33.2$\pm$4.5} & 32.9$\pm$2.0 & -0.3 [-4.8, 3.8] \\
\bottomrule
\end{tabular}
\caption{MCP-style ARC-Challenge scoring stress tests on DistilGPT-2. MCP label lists all choices and scores the next label token; MCP text lists all choices and scores each answer text. The mixed pattern is why the main scoring-protocol table focuses on CP text and first-token prefilling.}
\label{tab:mcp-stress-test}
\end{table}

\section{Backbone Transfer Details}
\label{app:backbone-transfer}

Table~\ref{tab:gpt2-arc} collects the same-family cross-backbone comparisons in one place.
It shows that the IA3-family improvement pattern extends beyond DistilGPT-2 to GPT-2 small, Pythia-410M, and Qwen2.5-1.5B.
At the same time, additive adapters show smaller or neutral effects, which is why the main empirical conclusion stays centered on multiplicative control rather than adapter-independent gains.
The table therefore acts as the compact backbone-transfer summary behind the scattered headline rows in the main text.

\begin{table*}[!t]
\centering
\scriptsize
\setlength{\tabcolsep}{3.0pt}
\begin{tabular}{@{}lllrrrrr@{}}
\toprule
Base adapter & Backbone & Same-family comparison & Params & Seeds & Base & PREFACE & $\Delta$ \\
\midrule
IA3 & DistilGPT-2 & IA3 $\rightarrow$ \smallmethod{} & 147,540 & 3 & 24.4$\pm$0.3 & \textbf{28.7$\pm$2.8} & +4.2 \\
LoRA & DistilGPT-2 & LoRA-tuned $\rightarrow$ \lorasidecar{} & 270,632 & 3 & 32.6$\pm$2.7 & \textbf{34.3$\pm$0.7} & +1.8 \\
DoRA & DistilGPT-2 & DoRA $\rightarrow$ \dorasidecar{} & 270,632 & 3 & 34.3$\pm$4.0 & \textbf{34.4$\pm$3.2} & +0.1 \\
IA3 & GPT-2 small & IA3 $\rightarrow$ \smallmethod{} & 239,700 & 5 & 24.5$\pm$1.5 & \textbf{29.0$\pm$2.3} & +4.5 \\
IA3 & GPT-2 medium & IA3 $\rightarrow$ \smallmethod{} & 565,332 & 3 & 32.2$\pm$1.0 & \textbf{33.6$\pm$5.7} & +1.3 \\
IA3 & Pythia-410M & IA3 $\rightarrow$ \gatedmethod{} & 467,052 & 3 & 31.5$\pm$1.5 & \textbf{34.0$\pm$0.8} & +2.5 \\
IA3 & Qwen2.5-0.5B & IA3 $\rightarrow$ \method{} & 554,604 & 3 & 40.6$\pm$1.0 & \textbf{41.8$\pm$1.2} & +1.2 \\
IA3 & Qwen2.5-1.5B & IA3 $\rightarrow$ \gatedmethod{} & 1,164,912 & 3 & 51.5$\pm$1.0 & \textbf{53.1$\pm$0.2} & +1.6 \\
LoRA & Qwen2.5-0.5B & LoRA-tuned $\rightarrow$ \lorasidecar{} & 87,704 & 3 & 27.8$\pm$1.2 & \textbf{28.1$\pm$1.5} & +0.3 \\
\bottomrule
\end{tabular}
\caption{ARC-Challenge validation accuracy by base adapter and backbone using separation-style continuation scoring. ``Params'' counts trainable parameters in the reported adaptation stage, including the static scale in jointly trained -small rows. Two-stage -gated rows freeze the static scale. Rows compare against same-family references rather than IA3 as a universal baseline.}
\label{tab:gpt2-arc}
\end{table*}

\section{Representation-Side Baseline}
\label{app:loreft-baseline}

Table~\ref{tab:loreft-related-baseline} summarizes the direct representation-side baselines discussed in the main text.
The table clarifies that PARA/PEDRO-style adapters are competitive, but the gains are smaller and less uniform than the multiplicative IA3-family result.
It also shows that the best raw accuracy in this appendix section comes from a representation-side adapter, not from the IA3 family, which keeps the paper from overclaiming universal dominance.
What remains distinctive for the main proposal is not winning every adapter family, but providing one bounded prefix controller across PEFT families and downstream inference formats; its strongest measured gains appear in multiplicative settings.

\begin{table*}[!t]
\centering
\scriptsize
\setlength{\tabcolsep}{5.0pt}
\begin{tabular}{@{}lrrL{6.0cm}@{}}
\toprule
Method & Params & ARC Acc. & Paired evidence \\
\midrule
IA3 & 18,432 & 24.4$\pm$0.3 & -- \\
LoReFT-minimal & 18,552 & 28.8$\pm$0.6 & -- \\
PREFACE-LoReFT & 14,004 & 29.0$\pm$1.3 & +0.2 [-0.9, 1.7] vs LoReFT \\
PARA/PEDRO-small & 60,056 & \secondbest{30.5$\pm$2.3} & -- \\
PREFACE-PARA/PEDRO-small & 13,908 & \textbf{31.5$\pm$1.7} & +1.0 [0.0, 2.5] vs PARA/PEDRO \\
\smallmethod{} & 147,540 & 28.7$\pm$2.8 & -1.9 [-7.9, 3.3] vs PARA/PEDRO \\
\bottomrule
\end{tabular}
\caption{Direct representation-side diagnostics on DistilGPT-2 ARC-Challenge over seeds 13, 21, and 42. Bold marks the best accuracy and underline marks the second-best accuracy. LoReFT-minimal is a compact local LoReFT-style reproduction with rank-4 low-rank subspace interventions on context-token hidden states. PREFACE-LoReFT freezes this base adapter and trains a prefix-conditioned controller over its rank channels. PARA/PEDRO-small uses prefix-pooled representation control and token--prompt dot-product gates to add low-rank hidden-state residuals. PREFACE-PARA/PEDRO-small freezes that base adapter and trains a prefix-conditioned controller over its value channels. Paired intervals are hierarchical bootstrap intervals in percentage points.}
\label{tab:loreft-related-baseline}
\end{table*}

\clearpage
\section{Controller Architecture Ablation}
\label{app:controller-ablation}

Table~\ref{tab:controller-ablation} details the controller ablation behind the TokenMLP choice.
The main takeaway is that lightweight token-local processing is sufficient, while stronger token mixing does not reliably improve accuracy.
TokenMLP gives the best compact result in this sweep, but MeanMLP and TokenCNN remain close, which suggests that performance is not overly sensitive to small controller-architecture changes.
By contrast, heavier attention-style token mixing increases parameters substantially without delivering a clear accuracy advantage.

\begin{table*}[!t]
\centering
\scriptsize
\setlength{\tabcolsep}{2.5pt}
\begin{tabular}{@{}lrrrl@{}}
\toprule
Controller & Params & Acc. & $\Delta$ vs IA3 & Summary path \\
\midrule
Static IA3 & 18k & 24.4$\pm$0.3 & -- & static scale \\
MeanMLP & 148k & \secondbest{28.7$\pm$2.8} & \secondbest{+4.2} & mean $\rightarrow$ MLP \\
TokenMLP & 148k & \textbf{29.2$\pm$2.5} & \textbf{+4.8} & shared MLP $\rightarrow$ mean \\
Attention pooling & 203k & 27.9$\pm$0.8 & +3.5 & learned weighted mean \\
Self-attention & 149k & 28.5$\pm$2.8 & +4.1 & self-attn $\rightarrow$ mean \\
TokenCNN & 240k & \secondbest{28.7$\pm$2.9} & \secondbest{+4.2} & Conv1D $\rightarrow$ mean \\
TokenSelfAtt & 37.9M & 28.1$\pm$1.2 & +3.7 & full token self-attn \\
GatedTokenMLP & 197k & 27.7$\pm$2.3 & +3.2 & token MLP + gate \\
\bottomrule
\end{tabular}
\caption{DistilGPT-2 ARC-Challenge controller ablation over seeds 13, 21, and 42 using separation-style scoring and current-prefix control. Bold marks the best score and underline marks tied second-best scores. $h_1,\ldots,h_T$ denote the prefix-token hidden states seen by the controller. MeanMLP first averages these states and then applies the controller MLP; TokenMLP applies a shared MLP to each token state and then averages. TokenMLP is the best compact variant here, but stronger token mixing with attention or CNN layers does not reliably improve accuracy.}
\label{tab:controller-ablation}
\end{table*}

\section{Paired Evidence Details}
\label{app:paired-analysis}

Table~\ref{tab:paired-analysis} reports the full paired-evidence breakdown supporting the headline rows.
It distinguishes intervals that exclude zero from intervals that overlap zero and makes the adapter-family-specific pattern explicit.
The cleanest evidence remains the IA3-family ARC-Challenge rows on DistilGPT-2, GPT-2 small, Pythia-410M, and Qwen2.5-1.5B, whereas SciQ, LoRA, and DoRA are weaker or null.
This table is therefore the main statistical evidence for the paper's conclusion that the controller effect is real but not universal.

\begin{table*}[!t]
\centering
\scriptsize
\setlength{\tabcolsep}{2.4pt}
\begin{tabular}{@{}llL{3.2cm}rrL{2.6cm}L{4.2cm}@{}}
\toprule
Family & Setting & Reference $\rightarrow$ prefix variant & Ref. & Prefix & $\Delta$ [95\%] & Seed-level evidence \\
\midrule
Mult. & SciQ & IA3 $\rightarrow$ \sharedmethod{} & 51.5 & \textbf{52.5} & +1.0 [-0.5, 2.5] & 2/3 seeds; boot $p=0.191$; McNemar diag. $p=0.172$ \\
Mult. & ARC & IA3 $\rightarrow$ \smallmethod{} & 24.4 & \textbf{28.7} & \textbf{+4.2 [0.1, 8.1]} & 3/3 seeds; boot $p=\textbf{0.046}$; McNemar diag. $p=0.0049$ \\
Mult. & WinoGrande & IA3 $\rightarrow$ \smallmethod{} & 48.8 & \textbf{49.8} & +1.0 [-0.6, 2.5] & 3/3 seeds; boot $p=0.205$; McNemar diag. $p=0.196$ \\
Mult. & ARC, GPT-2 & IA3 $\rightarrow$ \smallmethod{} & 24.5 & \textbf{29.0} & \textbf{+4.5 [1.0, 8.3]} & 5/5 seeds; boot $p=\textbf{0.0088}$; McNemar diag. $p=0.00085$ \\
Mult. & ARC, GPT-2-M & IA3 $\rightarrow$ \smallmethod{} & 32.2 & \textbf{33.6} & +1.3 [-4.3, 7.6] & 1/3 seeds; boot $p=0.686$; McNemar diag. $p=0.502$ \\
Mult. & ARC, Pythia & IA3 $\rightarrow$ \gatedmethod{} & 31.5 & \textbf{34.0} & \textbf{+2.5 [0.7, 4.2]} & 3/3 seeds; boot $p=\textbf{0.013}$; McNemar diag. $p=0.0032$ \\
\midrule
Mult. & ARC, Qwen & IA3 $\rightarrow$ \method{} & 40.6 & \textbf{41.8} & +1.2 [-0.2, 2.7] & 3/3 seeds; boot $p=0.101$ \\
Mult. & ARC, Qwen-1.5B & IA3 $\rightarrow$ \gatedmethod{} & 51.5 & \textbf{53.1} & \textbf{+1.6 [0.2, 3.0]} & 3/3 seeds; boot $p=\textbf{0.033}$ \\
Mult. & SciQ, Qwen & IA3 $\rightarrow$ \method{} & \textbf{81.7} & 81.4 & -0.3 [-0.8, 0.2] & 0/3 seeds; boot $p=0.298$ \\
\midrule
LoRA & ARC, sep. & LoRA-tuned $\rightarrow$ \lorasidecar{} & 32.6 & \textbf{34.3} & +1.8 [-1.6, 5.1] & 2/3 seeds; boot $p=0.294$; McNemar diag. $p=0.205$ \\
LoRA & ARC, Qwen & LoRA-tuned $\rightarrow$ \lorasidecar{} & 27.8 & \textbf{28.1} & +0.3 [-1.2, 1.9] & 1/3 seeds; boot $p=0.708$ \\
LoRA & SciQ, Qwen & LoRA-tuned $\rightarrow$ \lorasidecar{} & \textbf{75.0} & 75.0 & 0.0 [-0.1, 0.0] & 0/3 seeds; boot $p=0.981$ \\
DoRA & ARC, sep. & DoRA $\rightarrow$ \dorasidecar{} & 34.3 & \textbf{34.4} & +0.1 [-3.1, 3.5] & 1/3 seeds; boot $p=0.982$; McNemar diag. $p=1.000$ \\
\bottomrule
\end{tabular}
\caption{Seed-level paired evidence by adapter family on length-normalized accuracy. Deltas are candidate--reference accuracy in percentage points. The interval is a hierarchical bootstrap that resamples seeds and then examples within each seed; this is the primary uncertainty estimate. Pooled McNemar values, when shown, are ancillary diagnostics because seed--example pairs are not independent.}
\label{tab:paired-analysis}
\end{table*}

\section{Checkpoint and Runtime Details}
\label{app:checkpoint-details}

Tokenizers are loaded from the same checkpoint revisions as the corresponding backbones.

\begin{table*}[!htbp]
\centering
\scriptsize
\setlength{\tabcolsep}{4.0pt}
\begin{tabular}{@{}lll@{}}
\toprule
Model identifier & Revision & Precision \\
\midrule
\texttt{distilgpt2} & \texttt{2290a62682d06624634c1f46a6ad5be0f47f38aa} & float32 \\
\texttt{gpt2} & \texttt{607a30d783dfa663caf39e06633721c8d4cfcd7e} & float32 \\
\texttt{gpt2-medium} & \texttt{6dcaa7a952f72f9298047fd5137cd6e4f05f41da} & float32 \\
\texttt{EleutherAI/pythia-410m-deduped} & \texttt{c4fc8d586d62df497f1f9b69d66d3ca419992d3e} & float32 \\
\texttt{Qwen/Qwen2.5-0.5B-Instruct} & \texttt{7ae557604adf67be50417f59c2c2f167def9a775} & float16 \\
\texttt{Qwen/Qwen2.5-1.5B-Instruct} & \texttt{989aa7980e4cf806f80c7fef2b1adb7bc71aa306} & float16 \\
\texttt{Qwen/Qwen3.5-0.8B} & \texttt{2fc06364715b967f1860aea9cf38778875588b17} & float16 \\
\bottomrule
\end{tabular}
\caption{Exact checkpoint revisions recovered from the local Hugging Face cache metadata used by the reported runs.}
\label{tab:checkpoint-details}
\end{table*}

\section{Controller Overhead}
\label{app:sidecar-latency}

Table~\ref{tab:sidecar-latency} reports the latency cost of the controller relative to static adapters.
The overhead is measurable but remains small in absolute memory usage on the laptop-GPU setup.
In throughput terms, PREFACE variants are slower than static IA3, LoRA, or DoRA because they add a prefix encoder and channel-wise modulation.
Even so, the peak allocated memory stays below 0.4GB in this benchmark, which helps frame the method as lightweight in deployment rather than just in trainable parameters.

\begin{table}[!t]
\centering
\scriptsize
\setlength{\tabcolsep}{2.6pt}
\begin{tabular}{@{}lrrrrr@{}}
\toprule
Method & Params & ms/q & q/s & choices/s & Peak MB \\
\midrule
Frozen & 0 & 51.6 & 19.4 & 77.1 & 377.8 \\
IA3 & 18k & 52.5 & 19.0 & 75.7 & 377.9 \\
LoRA-tuned & 184k & 60.6 & 16.5 & 65.7 & 378.5 \\
DoRA & 189k & 58.0 & 17.2 & 68.6 & 378.5 \\
\smallmethod{} & 148k & 75.0 & 13.3 & 53.1 & 378.5 \\
\lorasidecar{} & 271k & 85.6 & 11.7 & 46.5 & 379.7 \\
\dorasidecar{} & 271k & 88.9 & 11.2 & 44.8 & 379.7 \\
\bottomrule
\end{tabular}
\caption{Runtime on 100 ARC-Challenge validation questions using DistilGPT-2 on the laptop GPU. Choices are scored sequentially and prefix representations are not cached. PREFACE adds a small prefix encoder and bounded channel-wise modulation, increasing latency relative to static adapters while remaining below 0.4GB peak allocated CUDA memory in this DistilGPT-2 benchmark.}
\label{tab:sidecar-latency}
\end{table}

\section{Prefix-Source Diagnostics}
\label{app:options-diagnostic}

Table~\ref{tab:qwen35-mechanism} gives a parameter-matched multi-task mechanism diagnostic on Qwen3.5-0.8B.
The learned-global control exactly matches IA3, while current, shuffled, and random representations differ by at most 0.19 pooled points and all current-minus-control intervals include zero.
No prefix-source contrast is significant before correction, so Holm correction does not change any conclusion.
The learned representations nevertheless retain coarse task information: a held-out nearest-centroid probe predicts task identity at 82.1\%, 80.2\%, and 83.7\% across seeds (33.3\% chance), with effective ranks 8.24, 8.30, and 15.38.
Thus the controller is non-constant and task-structured, but these controls do not establish fine-grained semantic matching to the current question.

\begin{table}[!htbp]
\centering
\scriptsize
\setlength{\tabcolsep}{3.5pt}
\begin{tabular}{@{}lrrl@{}}
\toprule
Control source & Pooled acc. & Current minus control & 95\% interval \\
\midrule
IA3 (no controller) & 67.74 & +0.11 & [-0.37, 0.59] \\
Learned global & 67.74 & +0.11 & [-0.37, 0.59] \\
Current prefix & 67.86 & -- & -- \\
Within-task shuffle & 67.82 & +0.04 & [-0.30, 0.37] \\
Cross-task shuffle & 67.74 & +0.11 & [-0.22, 0.45] \\
Random vector & 67.93 & -0.07 & [-0.48, 0.30] \\
\bottomrule
\end{tabular}
\caption{Matched Qwen3.5 mechanism diagnostic pooled over 299 fixed examples from each of ARC-Challenge, SciQ, and PIQA and three seeds. One controller is shared across tasks. Intervals hierarchically resample seeds and examples.}
\label{tab:qwen35-mechanism}
\end{table}

Table~\ref{tab:separation-prefix-diagnostic} gives the separation-style prefix-source diagnostic for all three adapter families.
It shows that IA3 and LoRA benefit more than DoRA, and that shuffled prefixes remain surprisingly competitive.
That pattern constrains the interpretation: the controller is not best understood as a pure donor-question semantic mechanism.
Instead, the evidence points to a bounded prefix-derived control channel whose gains partly survive even when the source prefix is perturbed.

\begin{table*}[!htbp]
\centering
\scriptsize
\setlength{\tabcolsep}{2.4pt}
\begin{tabular}{@{}lllrr>{\raggedright\arraybackslash}p{5.2cm}@{}}
\toprule
Family & Method & Prefix source & Params & Acc. & Seed-level evidence vs. same-family reference \\
\midrule
Mult. & IA3 & static & 18k & 24.4$\pm$0.3 & -- \\
Mult. & Static basis (92k) & learned global & 92k & 27.5$\pm$3.0 & +3.1 [-0.7, 6.6]; 2/3 seeds; boot $p=0.120$ \\
Mult. & \smallmethod{} & current prefix & 148k & \secondbest{28.7$\pm$2.8} & +4.2 [0.2, 8.0]; 3/3 seeds; boot $p=0.049$ \\
Mult. & \smallmethod{} & random prefix & 148k & 27.6$\pm$1.7 & +3.2 [0.0, 6.4]; 3/3 seeds; boot $p=0.054$ \\
Mult. & \smallmethod{} & shuffled prefix & 148k & \textbf{28.8$\pm$2.7} & +4.3 [0.4, 8.1]; 3/3 seeds; boot $p=0.042$ \\
\midrule
LoRA & LoRA-tuned & static & 184k & 32.6$\pm$2.7 & -- \\
LoRA & \lorasidecar{} & current prefix & 271k & \secondbest{34.3$\pm$0.7} & +1.8 [-1.6, 5.2]; 2/3 seeds; boot $p=0.319$ \\
LoRA & \lorasidecar{} & random prefix & 271k & 32.3$\pm$2.0 & -0.2 [-2.0, 1.7]; 1/3 seeds; boot $p=0.833$ \\
LoRA & \lorasidecar{} & shuffled prefix & 271k & \textbf{34.7$\pm$1.0} & +2.1 [-1.3, 5.6]; 2/3 seeds; boot $p=0.253$ \\
\midrule
DoRA & DoRA & static & 189k & \secondbest{34.3$\pm$4.0} & -- \\
DoRA & \dorasidecar{} & current prefix & 271k & \textbf{34.4$\pm$3.2} & +0.1 [-3.0, 3.6]; 2/3 seeds; boot $p=0.957$ \\
DoRA & \dorasidecar{} & random prefix & 271k & 33.7$\pm$1.6 & -0.7 [-3.2, 2.2]; 1/3 seeds; boot $p=0.653$ \\
DoRA & \dorasidecar{} & shuffled prefix & 271k & 34.2$\pm$3.0 & -0.1 [-3.2, 3.3]; 2/3 seeds; boot $p=0.989$ \\
\bottomrule
\end{tabular}
\caption{DistilGPT-2 ARC-Challenge separation-style prefix-source diagnostic over seeds 13, 21, and 42. Bold marks the best score and underline marks the second-best score within each adapter family. Rows compare within adapter family: multiplicative rows use IA3 as the reference, PREFACE-LoRA rows use tuned LoRA, and PREFACE-DoRA rows use tuned DoRA. The 92k static basis is not parameter-matched to the 148k controller; matched controls appear in Tables~\ref{tab:matched-mechanism}--\ref{tab:matched-other}. PREFACE-LoRA preserves the current-prefix improvement pattern, random prefix removes most of it, and shuffled prefix remains close to current, while PREFACE-DoRA is neutral.}
\label{tab:separation-prefix-diagnostic}
\end{table*}

Table~\ref{tab:arc-family-diagnostic} adds the options-format multiplicative diagnostic to show that the same static-versus-prefix pattern is not limited to one prompt format.
As in the separation-style table, the shuffled-prefix row remains close to the current-prefix row, while the learned global control already captures part of the gain.
This pattern shows that gains can survive a changed donor prefix; it does not isolate boundedness from other parameterization or optimization effects.

\begin{table*}[!t]
\centering
\scriptsize
\setlength{\tabcolsep}{2.6pt}
\begin{tabular}{@{}lllrrL{3.0cm}L{5.1cm}@{}}
\toprule
Family & Method & Prefix source & Params & Acc. & $\Delta$ & Diagnostic note \\
\midrule
Mult. & IA3 & static & 18k & 27.2$\pm$0.7 & -- & Static low-parameter multiplicative reference. \\
Mult. & Parameter-matched static & learned global & 147k & 29.7$\pm$2.5 & +2.45 [0.33, 4.57] & Adds scale capacity without input conditioning. \\
Mult. & \smallmethod{} & current prefix & 148k & \secondbest{30.5$\pm$1.6} & +3.34 [0.45, 6.24] & Prefix-controlled scale; +0.89 [-1.78, 3.57] vs. the parameter-matched static control. \\
Mult. & \smallmethod{} & random prefix & 148k & 28.4$\pm$0.3 & current +2.12 [-0.45, 4.68] & Replaces the context representation with fixed random vectors. \\
Mult. & \smallmethod{} & shuffled prefix & 148k & \textbf{30.9$\pm$1.6} & current -0.33 [-0.78, 0.00] & Uses donor-prefix summaries; not worse than current-prefix states. \\
\bottomrule
\end{tabular}
\caption{DistilGPT-2 ARC-Challenge options-format multiplicative diagnostic. Bold marks the best score and underline marks the second-best score. All rows use the options-text train/eval format and report seed mean $\pm$ standard deviation over seeds 13, 21, and 42. Bracketed values are seed-level bootstrap 95\% intervals in percentage points.}
\label{tab:arc-family-diagnostic}
\end{table*}

\section{Boundedness Mechanism Diagnostic}

Table~\ref{tab:boundedness-mechanism} tests the regularization induced by the bounded correction
$\delta(q)=\rho\tanh(Bc(q))$ while keeping the DistilGPT-2 backbone, frozen IA3 adapter, controller width, basis size, training data, and update budget fixed.
A very small bound ($\rho=0.05$) changes fewer predictions but also yields fewer net corrections.
The main setting ($\rho=0.20$) gives the best correction trade-off in this diagnostic.
A looser bound ($\rho=1.00$) produces larger corrections and more prediction flips without improving the net number of corrected decisions.
This pattern is consistent with moderate bounded control acting as a regularizer, although the overlapping accuracy intervals do not establish a uniquely optimal value of $\rho$.

\begin{table*}[!htbp]
\centering
\scriptsize
\setlength{\tabcolsep}{3.1pt}
\begin{tabular}{@{}rrrrrrrr@{}}
\toprule
$\rho$ & Mean $|\delta|$ & P95 $|\delta|$ & Sum $\Delta$ & Mean $\Delta$ & Sum net/flips & Mean net/flips & Seeds \\
\midrule
0.05 & 0.0297 & 0.0491 & +1.00 [-0.33, 2.45] & -0.11 [-1.45, 1.23] & +9 / 69 & -1 / 80 & 3 \\
0.20 & 0.0662 & 0.1776 & \textbf{+2.01} [0.00, 4.12] & \textbf{+2.01} [-2.01, 5.80] & \textbf{+18} / 147 & \textbf{+18} / 234 & 3 \\
1.00 & 0.1199 & 0.3852 & +1.78 [-0.56, 4.12] & +1.56 [-1.11, 4.46] & +16 / 185 & +14 / 312 & 3 \\
\bottomrule
\end{tabular}
\caption{Boundedness diagnostic on DistilGPT-2 ARC-Challenge. Accuracy deltas are percentage points relative to the same frozen IA3 adapter, with paired hierarchical 95\% bootstrap intervals. ``Net/flips'' is wrong-to-right minus right-to-wrong decisions, followed by the total number of changed predictions pooled across three seeds. Correction magnitudes are measured from actual controller outputs over the validation examples.}
\label{tab:boundedness-mechanism}
\end{table*}

\section{Matched-Study Diagnostics}
\label{app:matched-study}
Tables~\ref{tab:matched-mechanism} and~\ref{tab:matched-other} place the complete primary accuracy controls in the main paper.
The learned-global arm retains the complete controller parameterization but receives a constant input. Its nominal parameter count matches the dynamic arm; input variation and effective functional degrees of freedom do not. This is why we also use a fixed-mean intervention on the same trained dynamic checkpoint. That intervention requires no retraining and removes input variation without changing learned weights.
The random control matches marginal scale but not learned correlations. We therefore interpret it together with three donor shuffles, which preserve the empirical correction distribution. All reported outcomes are retained. Three seeds and the use of development data limit confirmatory interpretation.
The unbounded arm changes only the output nonlinearity: $\rho\tanh(v)$ becomes $\rho v$. It is an optimization-matched diagnostic, not a separately tuned competitor. A difference cannot establish that boundedness is universally necessary.

\section{Multiplicity and Frozen-Adapter Test Details}
\label{app:ac-statistics}

Table~\ref{tab:matched-other} separates pointwise and multiplicity-adjusted evidence. The 12-comparison sensitivity family includes six current--global and six current--frozen contrasts, not all paper comparisons. Table~\ref{tab:ac-multiplicity} supplies the current--frozen intervals omitted from that main-paper table. The 36-comparison sensitivity additionally includes current versus mean, shuffled, random, and unbounded corrections in all six cells. Each interval uses 200,000 paired seed/example draws, checked with a second bootstrap random seed. Monte Carlo precision does not resolve uncertainty from having only three training seeds.

\begin{table}[t]\centering\scriptsize
\setlength{\tabcolsep}{2.3pt}
\begin{tabular}{@{}llrr@{}}\toprule
Task & Adapter & Current $-$ frozen B12 & B36 \\
\midrule
ARC-C & IA3 & [-2.01, 9.36] & [-2.79, 10.03] \\
ARC-C & LoRA & [-3.46, 6.69] & [-4.24, 7.36] \\
ARC-C & DoRA & [-4.35, 8.36] & [-5.02, 9.36] \\
SciQ & IA3 & [-1.87, 4.80] & [-2.30, 5.30] \\
SciQ & LoRA & [-1.90, 4.40] & [-2.37, 4.90] \\
SciQ & DoRA & [0.00, 5.83] & [-0.40, 6.30] \\
\bottomrule\end{tabular}
\caption{Bonferroni sensitivity intervals in percentage points. The complete 36-contrast record is retained with raw predictions; these adjustments do not address development-set selection or every comparison in the paper.}
\label{tab:ac-multiplicity}
\end{table}

\paragraph{ARC test comparison.} After freezing the three current/global controller pairs used in the matched ARC validation diagnostic, we score all 1,172 official test examples with the same mean-answer-log-probability protocol, without retraining or test-driven selection. Both arms add 129,108 nominal parameters during training; base IA3 remains frozen. Recomputed zero-correction predictions agree with all 3,516 original baseline correctness flags. Table~\ref{tab:matched-other} reports the current--global result and B37 sensitivity (the earlier 36 diagnostic comparisons plus this test contrast). This does not change the jointly trained headline checkpoint or its 26.4\% accuracy.

\begin{table}[t]\centering\small
\setlength{\tabcolsep}{3.5pt}
\begin{tabular}{@{}llrr@{}}\toprule
Task & Adapter & Current bounded & Unbounded \\
\midrule
ARC-C & IA3 & \textbf{28.2$\pm$0.8} & 27.9$\pm$0.8 \\
ARC-C & LoRA & \textbf{34.1$\pm$1.2} & \textbf{34.1$\pm$1.5} \\
ARC-C & DoRA & \textbf{35.9$\pm$1.5} & 34.8$\pm$1.5 \\
SciQ & IA3 & \textbf{52.9$\pm$0.6} & 52.7$\pm$0.4 \\
SciQ & LoRA & \textbf{52.7$\pm$1.0} & 52.6$\pm$1.3 \\
SciQ & DoRA & 53.5$\pm$0.6 & \textbf{53.6$\pm$1.0} \\
\bottomrule\end{tabular}
\caption{Matched bounded/unbounded validation controls. Unbounded replaces $\rho\tanh(v)$ by $\rho v$ without retuning. Bold denotes the higher displayed mean, including rounding ties. All paired pointwise intervals for this contrast include zero.}
\label{tab:ac-unbounded}
\end{table}

\section{Exploratory Shared-Task Controller}
\label{app:heterogeneous}

We additionally test whether task heterogeneity makes actual-prefix information useful. One DistilGPT-2 IA3 adapter and one controller are shared across ARC-Challenge, SciQ, PIQA, and WinoGrande within each seed (13/21/42), with no task-specific adapter or head. We sample 1,024 training examples per task, train joint IA3 for 500 updates at learning rate $0.003$, then freeze it and train current/global controllers for 200 updates at $0.0005$. Every batch contains exactly one question per task. The matched 129,108-parameter architecture, initialization, objective, length-96 training limit, and controller batches are fixed across arms; no validation tuning is performed.

Within-task and cross-task interventions reuse complete donor correction bundles from the current checkpoint, with three fixed bijections and no self-donors; cross-task donors always come from another task. Bijections preserve the global correction marginal, although unequal task sizes preclude uniform donor-task frequencies. The mean intervention averages actual post-$\tanh$ corrections over 64 training prefixes per task. All evaluations use full validation splits and task-balanced macro averaging.

\begin{table*}[t]\centering\small
\setlength{\tabcolsep}{4.5pt}
\begin{tabular}{@{}lrrrrrr@{}}\toprule
Task & Frozen IA3 & Global & Current & Within shuffle & Cross shuffle & Train mean \\
\midrule
ARC-C & 21.63 & 22.41 & 23.63 & 23.60 & 23.52 & \textbf{23.97} \\
SciQ & 49.27 & \textbf{49.77} & 49.60 & 49.76 & 49.74 & \textbf{49.77} \\
PIQA & 59.07 & \textbf{59.18} & 58.92 & 58.93 & 58.94 & 58.89 \\
WinoGrande & 49.38 & 49.38 & 49.54 & \textbf{49.64} & 49.41 & 49.46 \\
Macro & 44.84 & 45.18 & 45.42 & 45.48 & 45.40 & \textbf{45.52} \\
\bottomrule\end{tabular}
\caption{Exploratory heterogeneous-task accuracy, mean over three seeds, one shared adapter/controller per seed. These checkpoints differ from all single-task headline rows. Bold marks highest displayed means, not statistical significance.}
\label{tab:heterogeneous}
\end{table*}

Macro current-minus-global is +0.24 points [-0.51, 0.94], current-minus-within-shuffle is -0.06 [-0.18, 0.09], and current-minus-cross-shuffle is +0.02 [-0.30, 0.35]. Current-minus-training-mean is -0.10 [-0.36, 0.11]. Intervals use 200,000 crossed seed/example draws, sharing seed resamples across tasks and resampling examples within each task. Neither task-level nor macro current--global/current--cross contrasts have strictly positive pointwise lower bounds. Bonferroni sensitivity across these ten primary contrasts also includes zero.

The macro input-varying correction RMS is 7.60\% of total residual-correction RMS, excluding identity and centering on the balanced training mean. Training-centroid task prediction from correction vectors reaches 63.8--72.9\% macro accuracy (chance 25\%), but this descriptive task information does not establish prediction utility. Relative to cross-task donors, current prefixes correct 0.42\% and harm 0.40\% of examples under macro averaging. Thus the experiment does not demonstrate either a current-prefix advantage or useful coarse task routing; similar accuracy is not an equivalence result. The joint base adapter is weaker than the separately trained headline models, and this conclusion is limited to the fixed short-budget diagnostic. The prespecified positive-evidence gate for Qwen replication was not met, so no outcome-driven Qwen search was run.

\section{Static-Compilation Protocol and Runtime}
\label{app:compilation}
The compilation audit uses every bounded-current checkpoint of the matched study: two tasks, three families, and three seeds. Corrections are the mean of saved post-$\tanh$ vectors over 256 training prefixes; no evaluation examples or labels enter their estimation. No training or hyperparameter search is performed. Full candidate scores and predictions are compared against the preserved fixed-mean intervention, with source hashes and per-example compiled outputs supplied in the software package.
GPT-2's implementation stores projection weights transposed relative to Eq.~\ref{eq:compilation}, so compilation multiplies the input-channel rows of the stored matrix. Bias is unchanged. Equivalence is algebraic in real arithmetic; float32 matrix reassociation can change argmax for near-tied answers. One such disagreement occurs in SciQ--IA3 seed 42. This is distinct from the statistically uncertain current--global contrast.
\paragraph{Timing protocol.} First 24 examples in preserved validation order; seed 13; float32, TF32 disabled, SDPA attention; four warmup questions per mode and five blocks in deterministically shuffled order. Each question packs all candidates into one batch. Dynamic mode computes a prefix-only pass, retains layer corrections on the GPU, then scores candidates. Fixed-unfolded uses the same training-mean vector without a prefix pass; compiled folds that vector into the weights; frozen merges the trained base adapter without correction. There is no KV-cache reuse in any arm. CUDA synchronization brackets each block. Tokenization, model loading, compilation, and file I/O are excluded. Each block's order, latency, throughput, and memory are retained.
\paragraph{Memory interpretation.} Model variants are kept resident while timing to avoid loading differences. We therefore report incremental peak allocated CUDA memory above the resident baseline, not an isolated process's total deployment memory. This includes transient tensors and does not estimate optimizer memory or multi-tenant storage. The same benchmark must be rerun for other decoding implementations or hardware.
\begin{table*}[t]\centering\scriptsize
\setlength{\tabcolsep}{5pt}
\begin{tabular}{@{}lllrrr@{}}\toprule
Task & Adapter & Mode & Latency median [min,max] ms & Questions/s & Incremental peak MiB \\\midrule
ARC-C & IA3 & frozen & 8.57 [8.43, 10.73] & 116.7 & 34.8 \\
ARC-C & IA3 & dynamic & 18.90 [18.35, 25.23] & 52.9 & 35.0 \\
ARC-C & IA3 & fixed-unfolded & 9.30 [9.22, 9.48] & 107.6 & 35.1 \\
ARC-C & IA3 & compiled & 8.43 [8.41, 10.93] & 118.6 & 34.8 \\
\addlinespace
ARC-C & LoRA & frozen & 8.47 [8.46, 13.64] & 118.0 & 34.8 \\
ARC-C & LoRA & dynamic & 18.29 [18.07, 18.61] & 54.7 & 35.0 \\
ARC-C & LoRA & fixed-unfolded & 9.14 [9.10, 9.29] & 109.4 & 35.1 \\
ARC-C & LoRA & compiled & 8.48 [8.38, 10.69] & 117.9 & 34.8 \\
\addlinespace
ARC-C & DoRA & frozen & 8.47 [8.34, 8.64] & 118.1 & 34.8 \\
ARC-C & DoRA & dynamic & 18.30 [18.18, 19.16] & 54.6 & 35.0 \\
ARC-C & DoRA & fixed-unfolded & 9.17 [9.14, 9.39] & 109.0 & 35.1 \\
ARC-C & DoRA & compiled & 8.44 [8.35, 8.60] & 118.5 & 34.8 \\
\addlinespace
SciQ & IA3 & frozen & 8.14 [8.11, 8.26] & 122.9 & 10.2 \\
SciQ & IA3 & dynamic & 18.16 [17.82, 18.47] & 55.1 & 10.3 \\
SciQ & IA3 & fixed-unfolded & 8.91 [8.81, 8.97] & 112.2 & 10.4 \\
SciQ & IA3 & compiled & 8.19 [8.12, 8.99] & 122.1 & 10.2 \\
\addlinespace
SciQ & LoRA & frozen & 8.17 [8.04, 8.21] & 122.4 & 10.2 \\
SciQ & LoRA & dynamic & 18.23 [18.14, 18.47] & 54.8 & 10.3 \\
SciQ & LoRA & fixed-unfolded & 8.93 [8.74, 9.01] & 111.9 & 10.4 \\
SciQ & LoRA & compiled & 8.19 [8.08, 8.28] & 122.1 & 10.2 \\
\addlinespace
SciQ & DoRA & frozen & 8.15 [8.07, 8.31] & 122.7 & 10.2 \\
SciQ & DoRA & dynamic & 18.25 [18.06, 18.46] & 54.8 & 10.3 \\
SciQ & DoRA & fixed-unfolded & 8.94 [8.88, 9.07] & 111.8 & 10.4 \\
SciQ & DoRA & compiled & 8.13 [8.06, 8.19] & 123.1 & 10.2 \\
\addlinespace
\bottomrule\end{tabular}
\caption{Compilation runtime audit. Block ranges describe measurement variability, not confidence intervals over tasks or training seeds. All rows use the same fixed protocol; earlier runtime tables use a different benchmark and should not be compared directly.}
\label{tab:folding-runtime}
\end{table*}

\section{Implementation and Reference Settings}
\label{app:implementation}
\subsection{Layer-Shared Input Controller}

The non-shared controller form learns a separate controller network for each adapted layer.
This improves flexibility but makes the controller much larger than a static vector adapter.
Most large sweeps use multiplicative PEFT, so the notation below writes the layerwise control as a channel scale $a_\ell$; additive and representation-side PEFT use the same sharing idea with their own channel values.
\sharedmethod{} shares the prefix encoder across all adapted MLP layers while keeping layer-specific static scale vectors and low-rank scaling bases:
\begin{equation}
    a_\ell(q_\ell) = a_{0,\ell} + \rho \tanh(B_\ell c_\theta(q_\ell)).
\end{equation}
Here $c_\theta$ is shared, while $a_{0,\ell}$ and $B_\ell$ are layer-specific.
This preserves per-layer adaptation capacity but amortizes the expensive prefix encoder.
In DistilGPT-2, the non-shared controller has 794k trainable parameters, whereas \sharedmethod{} uses 270k parameters with the same basis size and hidden width.
We also test \smallmethod{}, a compact version with basis size 4 and hidden width 16, using 148k trainable parameters.
For architecture-transfer diagnostics, we also test \gatedmethod{}, a conservative two-stage variant that first trains a static scale baseline, freezes the resulting scale, and then trains only a context-conditioned correction with a learned scalar gate per layer.
For this variant,
\begin{equation}
    a_\ell(q_\ell) = \hat{a}_{0,\ell} + g_\ell \rho \tanh(B_\ell c_\theta(q_\ell)),
\end{equation}
where $\hat{a}_{0,\ell}$ is the frozen static baseline and $g_\ell=\sigma(\gamma_\ell)$ is a learned scalar gate.
The gates are initialized small and regularized so the dynamic component must improve a strong static baseline rather than overwrite it.

\subsection{Spectral Diagnostic}

Let $W \in \mathbb{R}^{d_{\mathrm{out}} \times d_{\mathrm{in}}}$ be a pretrained MLP projection and let
\begin{equation}
    W \approx U_r \mathrm{diag}(\sigma_r) V_r^\top
\end{equation}
be a rank-$r$ truncated SVD.
We define $W_\perp = W - U_r \mathrm{diag}(\sigma_r)V_r^\top$.
\spectralmethod{} scales only the dominant singular channels:
\begin{equation}
    f(x; q) =
    W_\perp x +
    U_r \mathrm{diag}\left(\sigma_r \odot s(q)\right)V_r^\top x,
\end{equation}
where $q$ is a context-prefix summary and $s(q) \in \mathbb{R}^r$ is an input-conditioned scaling vector.
Equivalently,
\begin{equation}
    f(x; q) =
    Wx +
    U_r \mathrm{diag}\left(\sigma_r \odot (s(q)-\mathbf{1})\right)V_r^\top x.
\end{equation}
Thus the pretrained projection is recovered exactly when $s(q)=\mathbf{1}$.
We parameterize the spectral scaling as a bounded residual update around the identity:
\begin{equation}
    s(q) = \mathbf{1} + \rho \tanh(B c_\theta(q)),
\end{equation}
where $\rho$ is a small scale, $c_\theta(q) \in \mathbb{R}^m$ is predicted by a lightweight controller network, and $B \in \mathbb{R}^{r \times m}$ is a learned basis with $m \ll r$.
The spectral row is a diagnostic for whether singular channels are a good basis for prefix-conditioned control.
The earlier LoRA/DoRA implementations scale the input to the complete adapted projection and include a trainable static input scale; they do not isolate the low-rank update branch.

\subsection{Reference Experiment Settings}
The following settings describe the earlier sweeps, not the matched frozen-adapter study.

\paragraph{Models.}
We use GPT-2-style causal language models so experiments can run under a fixed 8GB GPU budget.
Most experiments use DistilGPT-2.
We also run targeted GPT-2 small, GPT-2 medium, and Pythia-410M experiments on ARC-Challenge and SciQ to test whether the main pattern survives larger backbones.
For modern evaluations, we use Qwen2.5-0.5B-Instruct, Qwen2.5-1.5B-Instruct, and Qwen3.5-0.8B \citep{qwen2026qwen35} under the same local MCQ scoring protocol.

\paragraph{Tasks.}
The primary reported experiment uses SciQ \citep{welbl2017sciq}.
We also include WinoGrande \citep{sakaguchi2021winogrande}, ARC-Challenge \citep{clark2018arc}, and PIQA \citep{bisk2020piqa}.
Choices are scored by continuation likelihood on held-out validation splits, reporting both summed and length-normalized mean log-probability.
Most rows use validation splits as reproducible labeled evaluation sets. After selecting the DistilGPT-2 configuration, we additionally evaluate the fixed IA3 and \smallmethod{} checkpoints on the untouched ARC-Challenge test split (1,172 examples), which was not used for architecture or hyperparameter selection.
To test beyond MCQ ranking, we also evaluate SciQ answer generation by teacher-forced answer-token NLL and perplexity.
For SciQ and ARC-Challenge, the separation-style template is \texttt{Question: \{question\}\textbackslash nAnswer:} followed by each candidate answer text.
For WinoGrande, the context is the sentence prefix before the blank, and each candidate continuation is \texttt{\{option\}\{suffix\}}.
In the options-based ARC diagnostic, all choices are shown as \texttt{Options:\textbackslash nA. ...} before \texttt{Answer:}, and candidate answer texts are scored after that marker.
We additionally report a post-hoc scoring-protocol diagnostic on ARC-Challenge: the main table includes the standard CP answer-text score and FTP+prefill, which scores only the first discriminative answer token after prefilling any token prefix shared by all answer choices.
Appendix Table~\ref{tab:mcp-stress-test} reports MCP-label and MCP-text stress tests, where all choices are shown before scoring the label token or answer text.

\paragraph{Baselines.}
We compare pretrained models with no adaptation, LoRA and DoRA on MLP projections, IA3 on MLP projections, PEFT prompt tuning, PEFT prefix tuning, static spectral scaling, compact LoReFT- and PARA/PEDRO-style references, PREFACE-LoReFT, PREFACE-PARA/PEDRO-small, \spectralmethod{}, \method{}, \lorasidecar{}, and \dorasidecar{} as prefix-conditioned controller diagnostics.
The SciQ study includes full-validation results for LoRA, DoRA, and the strongest multiplicative variants.
The LoReFT-style diagnostic uses rank-4 low-rank subspace interventions on GPT-2 block outputs, masks interventions to context tokens under MCQ scoring, and trains 18.6k parameters, closely matching IA3's parameter count.
PREFACE-LoReFT freezes this LoReFT adapter and trains a 14.0k-parameter prefix controller over its rank-channel intervention.
The PARA/PEDRO-style reference uses a prefix-pooled representation controller and token--prompt dot-product gates to add low-rank hidden-state residuals, training 60.1k parameters.
PREFACE-PARA/PEDRO-small freezes that adapter and trains a 13.9k-parameter controller over its value channels.
Prompt tuning and prefix tuning are trained with the same MCQ contrastive objective and candidate-scoring pipeline as the adapter rows, providing a prompt-side PEFT comparison in the same code path.
Together with parameter-matched static, learned-global, and random/shuffled prefix-source controls, these rows are our matched in-code comparison to nearby input-conditioned and hypernetwork-style PEFT ideas.
Transformer-Squared and In-Place TTT are discussed above at the design level; reproducing them would require separate implementations and broader benchmark protocols, so our empirical comparisons are restricted to baselines run in the same code path.

\paragraph{Implementation.}
Adapters target the MLP output projection: \texttt{c\_proj} for GPT-2-style models and \texttt{dense\_4h\_to\_h} for GPT-NeoX/Pythia.
For input-conditioned adapters, the controller summary is masked to the context prefix during both training and evaluation, so candidate answer tokens never contribute to the adapter state used for continuation scoring.
The backbone is frozen.
In jointly trained PREFACE-IA3, including the main DistilGPT-2 small variant, the static scale $u_{\mathrm{id}}=a_{0,\ell}$ is optimized with the controller; in two-stage gated runs, the trained IA3 scale is frozen before controller training. LoRA, DoRA, and representation-side base adapters are likewise frozen, with $u_{\mathrm{id}}=\mathbf{1}$.
Adapters are trained for 500 steps on up to 2,048 training examples from each dataset.
IA3 and \method{}-family adapters use AdamW with learning rate $3\times10^{-3}$, batch size 4, max sequence length 96, and gradient clipping at 1.0.
The initial LoRA row uses rank 8, alpha 16, and learning rate $3\times10^{-3}$.
To avoid under-tuning LoRA, we run a single-seed validation sweep over ranks $\{4,8,16\}$ and learning rates $\{10^{-4},3\times10^{-4},10^{-3},3\times10^{-3}\}$ with $\alpha=2r$, then rerun the best settings over three seeds. Because selection and reporting use the same validation split, these LoRA rows are development results rather than confirmatory estimates.
For additive adapters, \lorasidecar{} freezes tuned LoRA and trains only the prefix controller; \dorasidecar{} freezes tuned DoRA and trains the same controller.
For the options-based ARC diagnostic, we also include a parameter-matched static control with 147k parameters: it uses IA3 plus learned layer-specific scaling bases, but replaces the prefix encoder with learned global coefficients.
DoRA uses rank 8, alpha 16, and learning rate $10^{-3}$ after a small stability sweep.
Prompt tuning and prefix tuning use 20 virtual tokens.
Prompt tuning uses learning rate $3\times10^{-2}$.
Prefix tuning uses projection, encoder hidden size 128, and learning rate $5\times10^{-3}$.
The answer-generation evaluation uses the same DistilGPT-2 IA3 and \smallmethod{} adapters, 200 training steps, and full SciQ validation.
The Qwen3.5 study fixes one configuration across ARC-Challenge, SciQ, and PIQA: 1,024 training examples, 200 updates, batch size 1, maximum length 96, and float16 precision.
All Qwen3.5 rows use seeds 13, 21, and 42 and the complete local validation splits (299 ARC-Challenge, 1,000 SciQ, and 1,838 PIQA examples).
Experiments use PyTorch 2.11.0 with CUDA 12.8, Transformers 5.5.4, and PEFT 0.18.1; GPT-style runs use float32, and no run uses gradient accumulation.
Headline PREFACE checkpoints use mean pooling followed by the controller MLP; the token-mixing sweep in Appendix Table~\ref{tab:controller-ablation} is post-hoc and did not select these checkpoints.
We report full validation results averaged over seeds 13, 21, and 42 unless otherwise noted.
The GPT-2 small ARC-Challenge study uses five seeds: 7, 13, 21, 42, and 99.
ARC-Challenge uses all 1,119 training examples, with no rows discarded during preprocessing.
The study is designed as a controlled efficiency evaluation rather than a broad benchmark comparison.

\begin{table}[t]
\centering
\scriptsize
\setlength{\tabcolsep}{1.2pt}
\begin{tabular}{@{}L{1.55cm}R{1.08cm}R{1.0cm}R{1.0cm}L{1.8cm}@{}}
\toprule
Method & Params & ARC-C & SciQ & Note \\
\midrule
IA3 & 18,432 & 24.4$\pm$0.3 & \secondbest{51.5$\pm$0.9} & static scale \\
Prompt tuning & 15,360 & 27.4$\pm$2.3 & 49.0$\pm$0.3 & 20 virtual tokens \\
Prefix tuning & 1,302,656 & \textbf{31.7$\pm$2.5} & 32.3$\pm$0.6 & projected prefix \\
\smallmethod{} & 147,540 & \secondbest{28.7$\pm$2.8} & \textbf{52.6$\pm$0.5} & IA3 + prefix controller \\
\bottomrule
\end{tabular}
\caption{Prompt-side PEFT baselines on DistilGPT-2 under the same training and scoring pipeline. Bold marks the best and underline the second-best score. Prefix tuning is strongest on ARC-Challenge at 1.3M parameters; \smallmethod{} is strongest on SciQ with fewer parameters.}
\label{tab:prompt-side-baselines}
\end{table}

\end{document}